\documentclass[sigconf]{acmart}
\usepackage{enumitem}
\usepackage{flushend}
\usepackage{booktabs}
\usepackage{todonotes}
\usepackage{epsfig}
\usepackage{graphicx}
\usepackage{latexsym}
\usepackage{amssymb}
\usepackage{lipsum}
\usepackage{subfig}
\usepackage{color}

\usepackage[normalem]{ulem}
\usepackage{bigstrut}
\usepackage{multirow}
\usepackage{siunitx}
\usepackage{ulem}
\usepackage{listings}
\usepackage{eso-pic}
\usepackage{tikz}
\usepackage{algorithm}
\usepackage{fancyhdr}
\usepackage[noend]{algpseudocode}
\usepackage{soul}
\usepackage{amsmath}
\usepackage{xspace}
\usepackage[toc,page]{appendix}
\usepackage{mathtools}
\usepackage[ruled,vlined,algo2e]{algorithm2e}
\usepackage[toc,page]{appendix}
\usepackage[compact]{titlesec}

\AtBeginDocument{%
  \providecommand\BibTeX{{%
    \normalfont B\kern-0.5em{\scshape i\kern-0.25em b}\kern-0.8em\TeX}}}

\copyrightyear{2019} 
\acmYear{2019} 
\acmConference[MICRO-52]{The 52nd Annual IEEE/ACM International Symposium on Microarchitecture}{October 12--16, 2019}{Columbus, OH, USA}
\acmBooktitle{The 52nd Annual IEEE/ACM International Symposium on Microarchitecture (MICRO-52), October 12--16, 2019, Columbus, OH, USA}
\acmPrice{15.00}
\acmDOI{10.1145/3352460.3358253}
\acmISBN{978-1-4503-6938-1/19/10}

\graphicspath{{figs/}}

\begin{document}


\title{ASV: Accelerated Stereo Vision System}

\date{}
\author{Yu Feng}
\email{yfeng28@ur.rochester.edu}
\affiliation{
    \institution{University of Rochester}
}

\author{Paul Whatmough}
\email{paul.whatmough@arm.com}
\affiliation{
\institution{Arm Research}
}

\author{Yuhao Zhu}
\email{yzhu@rochester.edu}
\affiliation{
\institution{University of Rochester}
}

\author{}
\affiliation{
\institution{\texttt{http://horizon-lab.org}}\vspace{20pt}}

\thispagestyle{firstpage}
\pagestyle{empty}

\setlength{\textfloatsep}{6pt}
\setlength{\floatsep}{6pt}

\titlespacing*{\section}{0pt}{8pt plus 0pt minus 0pt}{4pt plus 0pt minus 0pt}
\titlespacing*{\subsection}{0pt}{6pt plus 0pt minus 0pt}{3pt plus 0pt minus 0pt}



\newcommand{\website}[1]{{\tt #1}}
\newcommand{\program}[1]{{\tt #1}}
\newcommand{\benchmark}[1]{{\it #1}}
\newcommand{\fixme}[1]{{\textcolor{red}{\textit{#1}}}}

\newcommand*\circled[2]{\tikz[baseline=(char.base)]{
            \node[shape=circle,fill=black,inner sep=1pt] (char) {\textcolor{#1}{{\footnotesize #2}}};}}

\ifx\figurename\undefined \def\figurename{Figure}\fi
\renewcommand{\figurename}{Fig.}
\renewcommand{\paragraph}[1]{\textbf{#1}~~}
\newcommand{\figline}{{\vspace*{.05in}\hline}}

\newcommand{\Alg}[1]{Alg.~\ref{#1}}
\newcommand{\Sect}[1]{Sec.~\ref{#1}}
\newcommand{\Fig}[1]{Fig.~\ref{#1}}
\newcommand{\Tbl}[1]{Tbl.~\ref{#1}}
\newcommand{\Equ}[1]{Equ.~\ref{#1}}
\newcommand{\Apx}[1]{Appendix~\ref{#1}}

\newcommand{\specialcell}[2][c]{\begin{tabular}[#1]{@{}l@{}}#2\end{tabular}}
\newcommand{\note}[1]{\textcolor{red}{#1}}

\newcommand{\greenweb}{{\fontfamily{cmtt}\selectfont GreenWeb}\xspace}
\newcommand{\autogreen}{\textsc{AutoGreen}\xspace}
\newcommand{\proj}{\textsc{ASV}\xspace}
\newcommand{\mode}[1]{\underline{\textsc{#1}}\xspace}

\newcommand{\floor}[1]{\left\lfloor #1 \right\rfloor}
\newcommand{\ceil}[1]{\left\lceil #1 \right\rceil}

\newcommand{\RNum}[1]{\uppercase\expandafter{\romannumeral #1\relax}}


\begin{abstract}

Estimating depth from stereo vision cameras, i.e., "depth from stereo", is critical to emerging intelligent applications deployed in energy- and performance-constrained devices, such as augmented reality headsets and mobile autonomous robots. 
While existing stereo vision systems make trade-offs between accuracy, performance and energy-efficiency, we describe \proj, an accelerated stereo vision system that simultaneously improves both performance and energy-efficiency while achieving high accuracy.


The key to \proj is to exploit unique characteristics inherent to stereo vision, and apply stereo-specific optimizations, both algorithmically and computationally. We make two contributions. Firstly, we propose a new stereo algorithm, \textit{invariant-based stereo matching} (ISM), that achieves significant speedup while retaining high accuracy. The algorithm combines classic ``hand-crafted'' stereo algorithms with recent developments in Deep Neural Networks (DNNs), by leveraging the \textit{correspondence invariant} unique to stereo vision systems. 
Secondly, we observe that the bottleneck of the ISM algorithm is the DNN inference, and in particular the \textit{deconvolution} operations that introduce massive compute-inefficiencies. 
We propose a set of software optimizations that mitigate these inefficiencies. 
We show that with less than 0.5\% hardware area overhead, these algorithmic and computational optimizations can be effectively integrated within a conventional DNN accelerator. Overall, \proj achieves 5$\times$ speedup and 85\% energy saving with 0.02\% accuracy loss compared to today's DNN-based stereo vision systems.

\end{abstract}



\artifacts{ISM Algorithm: \url{https://github.com/horizon-research/ism-algorithm}\\Systolic-array Data-flow Optimizer: \url{https://github.com/horizon-research/systolic-array-dataflow-optimizer}}

\maketitle

\section{Introduction}
\label{sec:intro}

The demand for intelligent applications running on a diverse range of mobile and embedded platforms, such as micro-robots, augmented reality headsets, and smart-city sensor nodes, shows no sign of slowing down. 
A key primitive in these applications is estimating \textit{depth} information from the environment, which in turn serves as the building block for extracting higher-level semantics. 
For instance, depth information enables a mobile robot to detect and manipulate objects that are in close proximity.

Among numerous depth sensing techniques, we focus on \textit{stereo} camera systems, which estimate depth from a pair of horizontally displaced cameras that capture two different views of the scene, mimicking the human binocular vision. 
Compared to other depth sensing techniques such as LiDAR and structured-light sensors~\cite{wang20123d, weitkamp2006lidar}, stereo vision systems are much cheaper, consume less power, and are physically more compact~\cite{camera_comparison}. 
In response to the rising significance of stereo vision, recent mobile vision platforms integrate specialized stereo vision accelerators, such as the Stereo Depth Block in the Movidius Enhanced Vision Accelerator Suite~\cite{myriadx} and the Stereo \& Optical Flow Engine (SOFE) in the Nvidia Xavier mobile Systems-on-a-chip (SoC)~\cite{xaviersoc}.



\begin{figure}[t]
\centering
\includegraphics[width=0.7\columnwidth]{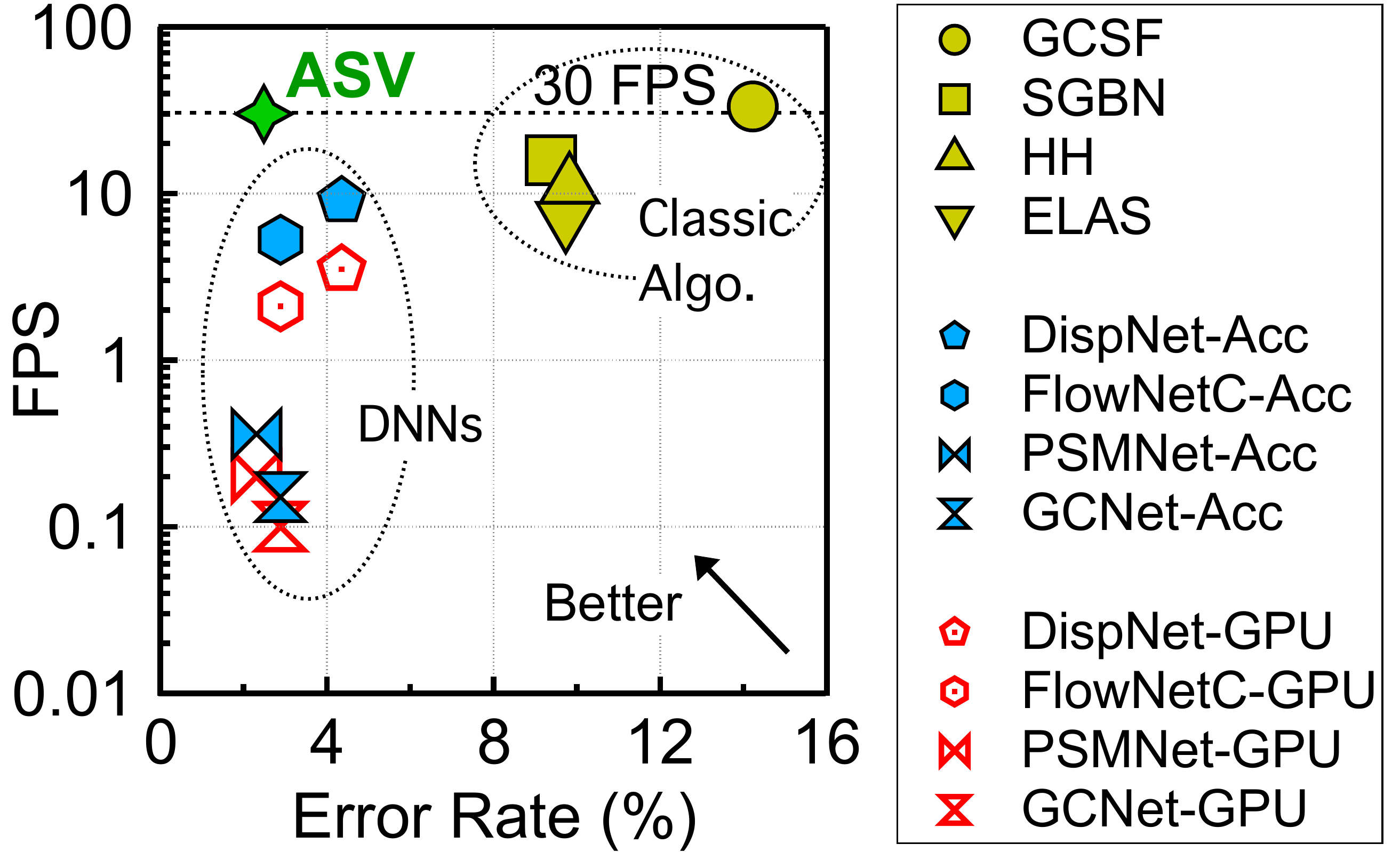}
\caption{\proj demonstrates both real-time (30 FPS) frame rates and DNN-like accuracy for stereo vision.}
\label{fig:trade-off}
\end{figure}

Stereo vision algorithms presented to date broadly define a frontier in the accuracy-efficiency design space.
\mbox{\Fig{fig:trade-off}} compares the frame rate and accuracy for four well-known classic stereo algorithms that use ``hand-crafted'' features, including GCSF~\mbox{\cite{Cech-CVPR-2011}}, SGBN~\mbox{\cite{SGBM2008HH}}, HH~\mbox{\cite{SGBM2008HH}}, and ELAS~\mbox{\cite{Elas2010Geiger}}, as well as four state-of-the-art DNNs solutions~\mbox{\cite{PSMNet2018, NvStereo2018, DispNet2015, GCNet2017}}. 
The DNN data is characterized on both a Pascal mobile GPU~\mbox{\cite{tx2dev}} (``-GPU'' suffix), as well as on a DNN accelerator~\mbox{\cite{samajdar2018scale}} (``-Acc'' suffix). 
In using low-dimensional ``hand-crafted'' features, classic algorithms lead to high error rates ($x$-axis), but are compute efficient, mostly operating at close to real-time (e.g., 30~FPS, $y$-axis).
In contrast, DNNs models achieve very low error rates, but require 2--5 orders of magnitude more arithmetic operations, resulting in much lower frame rates.





This paper presents \proj, an accelerated stereo vision system that operates in real-time while achieving DNN comparable accuracy. 
While today's vision accelerators are primarily built for monocular vision tasks, \proj exploits unique characteristics of stereo vision, and applies stereo-specific optimizations, both algorithmic and computational. 
Critically, we show that with careful algorithmic choices, these stereo-specific optimizations can largely be implemented on the same computer architecture as conventional DNN accelerators with a set of basic, yet principled, hardware extensions, which widens the applicability of this work.



At the core of \proj is a new low-latency, high-accuracy stereo vision algorithm. 
We exploit the temporal invariance introduced by stereo cameras: a single physical point projects to a unique pair of pixels on the left and right image planes; although the pixel locations move over time, their corresponding geometric relationship is fixed.
Our algorithm, ISM, uses compute-intensive DNNs to extract pixel correspondences from a small set of key frames. 
The correspondences are then propagated as initial estimates for subsequent non-key frames, where we make use of cheaper, classic algorithms. 
By combining learnt features from DNNs, and classic algorithms that explicitly model the physical world, ISM achieves high accuracy while reducing the compute cost.

While our ISM algorithm reduces the compute overhead, DNNs remain critical, as they generate the initial estimate of the correspondence information. 
We observe that stereo DNNs make heavy use of the \textit{deconvolution} operation\footnote{Deconvolution in deep learning is an incredibly unfortunate misnomer that should really be called ``transposed convolution.''} that exposes specific kernel sparsity, making conventional DNN accelerators inefficient. 
While prior work proposed specialized hardware to exploit deconvolution sparsity~\mbox{\cite{song2018GAN,Yazdanbakhsh2018GAN}}, we demonstrate that static software optimizations achieve better results without unnecessary hardware modifications.

Our approach is to transform an inherently sparse deconvolution layer into a sequence of dense convolutions, which can then be executed by canonical DNN accelerators. 
More importantly, this transformation uniquely exposes a new data reuse opportunity: \textit{inter-layer activation reuse} (ILAR), which does not exist in conventional DNNs.
While exhaustive search has been previously used to optimize data reuse patterns, it does not scale to optimizing deconvolution, because the transformation increases the layer count by up to 8$\times$, and ILAR adds another search dimension. 
Instead, we propose a constrained-optimization formulation, and demonstrate an efficient solver using dynamic programming.

We implement a software/hardware co-designed prototype of \proj. 
The hardware builds on top of a conventional systolic DNN accelerator~\cite{jouppi2017datacenter} implemented in 16nm technology. The \proj hardware minimally extends the baseline accelerator with less than 0.5\% area overhead. 
The software integrates the ISM algorithm and the deconvolution optimizations.

We evaluate \proj on a set of standard stereo vision benchmarks. Compared to the DNN baseline, \proj achieves 5$\times$ speedup and 85\% energy saving with 0.02\% accuracy loss. We also demonstrate the general applicability of software deconvolution, by applying it to Generative Adversarial Networks (GANs), which also make heavy use of deconvolutions. Under the same compute and memory resource constraints, we achieve 1.4 $\times$ speedup over a purpose-built deconvolution accelerator, due to the unique ILAR that we exploit.

To our best knowledge, this is the first paper that demonstrates a cost-effective stereo vision system. 
Using a software-hardware co-design approach, we show that carefully designed software optimizations achieve significant performance and energy improvements with simple, principled changes to existing DNN accelerators, which widens the applicability of our work. More specifically:

\begin{itemize}[topsep=0pt]
\item We propose the first stereo vision algorithm, ISM, that exploits \textit{temporal invariance} in stereo imaging to improve the performance with minimal accuracy loss;
\item We propose the first \textit{static} optimization framework for deconvolution, a key operation in stereo DNNs, which eliminates the sparsity-induced compute inefficiencies in deconvolution layers without hardware changes;
\item We are the first to identify \textit{inter-layer activation reuse} in deconvolution, \textit{a unique data reuse opportunity} exposed by our transformation framework, and which we exploit using an efficient constrained optimizer.
\item We co-design the hardware with the proposed software optimizations to achieve fast, low-power stereo vision \textit{with minimal changes to existing DNN accelerators}.
\end{itemize}

The remainder of the paper is organized as follows.~\Sect{sec:bcg} gives an overview of necessary background.
\Sect{sec:algo} introduces our invariant-based stereo matching algorithm.
\Sect{sec:deconv} describes the software optimizations for efficient implementation of the deconvolution operation.
\Sect{sec:sys} presents the design of \proj, including both software and hardware considerations.
\Sect{sec:exp} and \Sect{sec:eval} are experimental methodology and results, respectively.
\Sect{sec:related} positions \proj in the context of related work, and \Sect{sec:conc} concludes the paper.

\section{Background}
\label{sec:bcg}

We first describe the scope of our work: vision-based systems that extract 3D information from 2D stereo images (\Sect{sec:bcg:depth}). 
We then introduce the necessary background of stereo vision algorithms, including both classic hand-crafted algorithms and contemporary stereo DNNs (\Sect{sec:bcg:sv}).


\subsection{Depth Sensing}
\label{sec:bcg:depth}

There are two essential methods to extract depth information: passive sensing and active sensing. 
Passive sensing techniques observe the environment, primarily through cameras, and infer depth using computer vision algorithms. 
In contrast, active sensing techniques transmit signals and analyze the response to calculate depth; examples include structured light~\cite{wang20123d} and LiDAR~\cite{weitkamp2006lidar}.

This paper focuses on camera-based passive sensing. Compared to alternatives such as LiDAR, cameras are much cheaper and less bulky~\cite{camera_comparison}. As a result, camera-based depth sensing is widely adopted in systems such as autonomous vehicles and AR headsets. According to Allied Market Research, the adoption of stereo cameras is expected to grow 60.4\% by 2020~\cite{3dcammarket}. The recent industry trend of integrating dedicated stereo vision accelerators into mobile SoCs (e.g., Movidius~\cite{myriadx} and Nvidia~\cite{xaviersoc}) further underlines the significance of stereo vision for depth sensing.

\subsection{Depth From Stereo}
\label{sec:bcg:sv}

\begin{figure}[t]
\centering
\subfloat[\small{Triangulation.}]
{
  \includegraphics[trim=0 0 0 0, clip, height=1.65in]{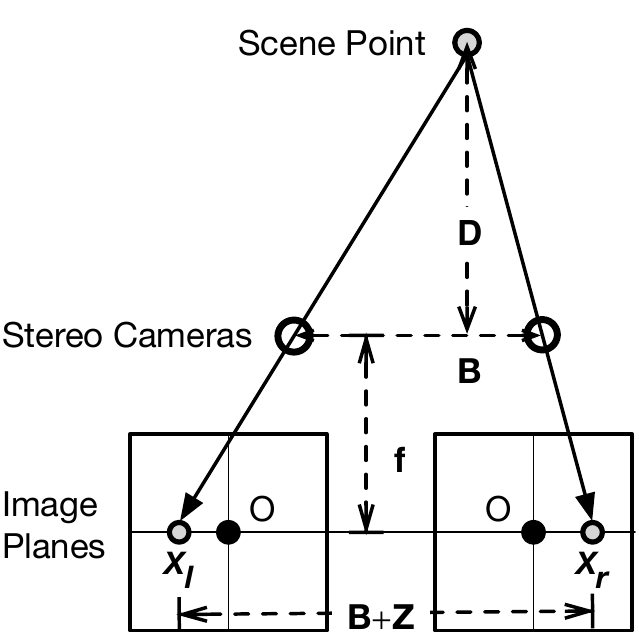}
  \label{fig:triangulation}
}
\subfloat[\small{Disparity Map.}]
{
  \includegraphics[trim=0 0 0 0, clip, height=1.65in]{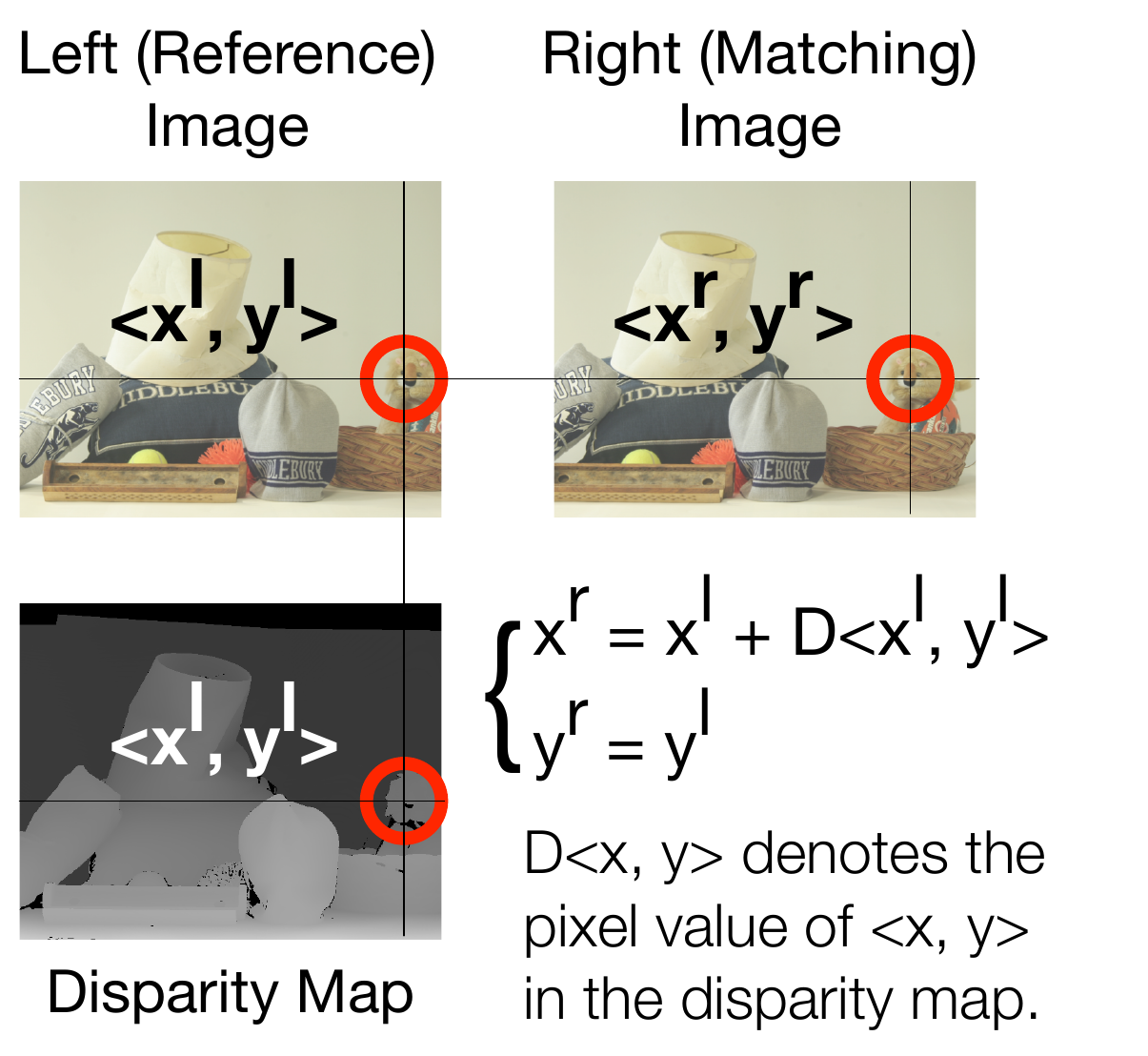}
  \label{fig:dispmap}
}
\caption{``Depth from stereo'' illustration: given an image pair, \textit{stereo matching} algorithms first generate the disparity map~\protect\subref{fig:dispmap}, from which depth is then calculated through \textit{triangulation}~\protect\subref{fig:triangulation}. Triangulation is computationally trivial; this paper focuses on optimizing stereo matching algorithms.}
\label{fig:sm}
\end{figure}


\paragraph{Triangulation} The key idea behind stereo depth estimation is that a single physical scene point projects to a unique pair of pixels, via two observing cameras; the horizontal displacement between the two pixels captured on the left and right image planes is inversely proportional to the distance of the point from the observer (i.e., the depth).~\Fig{fig:triangulation} illustrates this process, where the scene point is captured at position $x^l$ and $x^r$ on the left and right image planes, respectively. Using similar triangles, the depth $D$ is calculated by:
\begin{equation}
  \label{eq:triangulation}
  D = B f / Z,	
\end{equation}
\noindent where $f$ is the focal length of the cameras, $B$ is the distance between the two camera lenses, and $Z$ is the \textit{disparity} $x^r - x^l$, i.e., the horizontal displacement between the two corresponding pixels in the left and right images. This process is widely known as \textit{triangulation}~\cite{hartley2003multiple, szeliski2010computer}.



%

\paragraph{Stereo Matching and Disparity Map} Since both $B$ and $f$ are camera intrinsic parameters, the key to triangulation is to calculate the disparity $Z$. Given the left (reference) image and the right (matching) image, we must find the pixels in each image that are the projections of the same physical point, a process also known as \textit{stereo matching}. 
In the end, stereo matching generates a ``disparity map'', whose <$x, y$> coordinates are taken to be coincident with the pixel coordinates of the reference image.~\Fig{fig:dispmap} shows one such example, in which the correspondence between a pixel <$x^l, y^l$> in the left image and a pixel <$x^r, y^r$> in the right image is given by:
\begin{equation}
  \label{eq:dispdef}
  x^r = x^l + D^{\text{<}x^l, y^l\text{>}},~~~y^r = y^l,
\end{equation}
\noindent where D<$x^l, y^l$> denotes the pixel value at <$x^l, y^l$> in the disparity map. Note that the compute cost of triangulation is trivial (\Equ{eq:triangulation}), and thus we focus on stereo matching.


Stereo matching algorithms consist of three stages~\cite{stereo2002scharstein, stereo2003Brown}: Feature Extraction (FE), Matching Optimization (MO), and Disparity Refinement (DR). 
Both conventional algorithms~\cite{stereo2002scharstein, stereo2003Brown, SGBM2008HH, StereoClass2008} and DNN-based algorithms~\cite{StereoClass2008, PSMNet2018, NvStereo2018, DispNet2015, GCNet2017} follow this processing pipeline, but differ in their implementations. 
Conventional methods extract hand-crafted features (e.g., SIFT~\cite{sift}, HOG~\cite{hog}, or plain pixel values) in the FE stage, search the feature space of both images to find the matching pixels in the MO stage, and improve the disparity resolution in the DR stage using techniques such as iterative gradient descent~\cite{tian1986algorithms}. DNNs, in contrast, implement each stage using a set of learnt parameters.

\begin{figure}[t]
  \begin{minipage}[t]{0.45\columnwidth}
    \centering
    \includegraphics[trim=0 0 0 0, clip, height=1.55in]{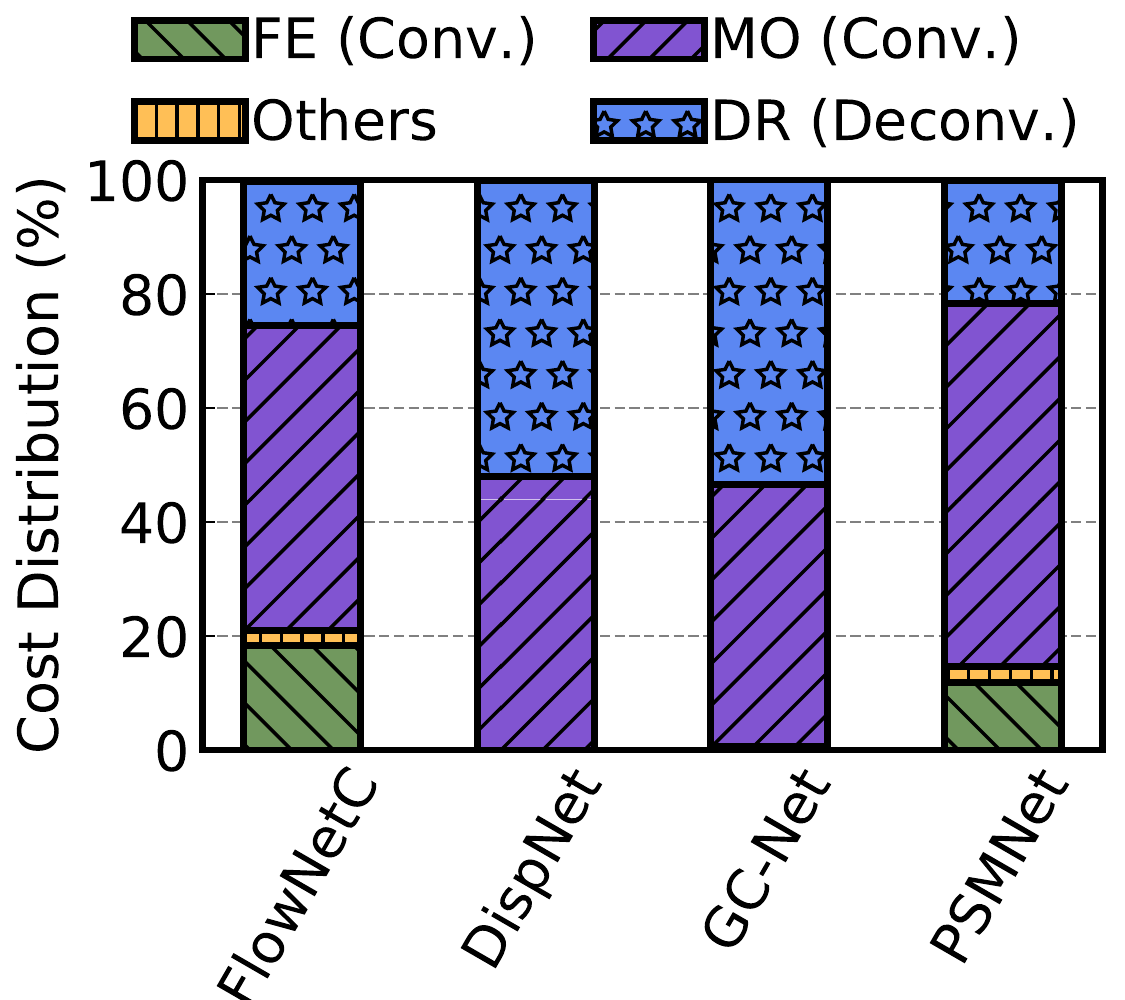}
    \caption{The arithmetic operation distribution of stereo matching DNNs.}
    \label{fig:cost_distribution}
  \end{minipage}
  \hspace{8pt}
  \begin{minipage}[t]{0.45\columnwidth}
    \centering
    \includegraphics[trim=0 0 0 0, clip, height=1.55in]{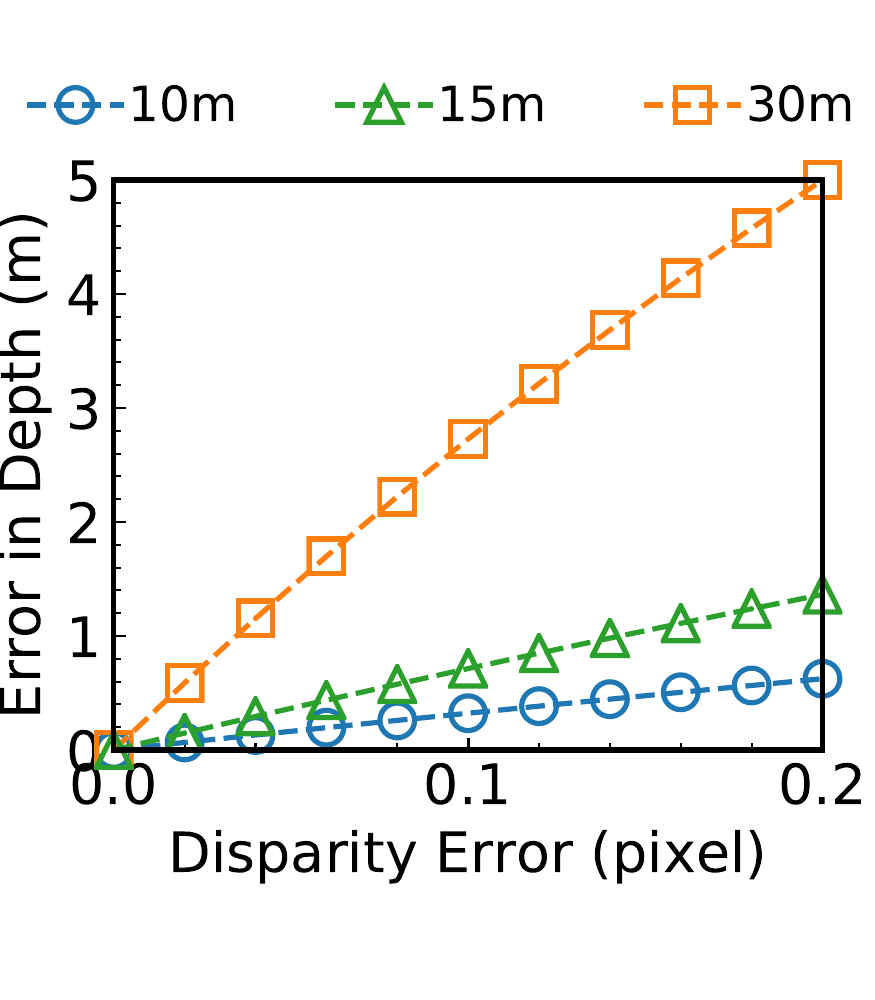}
    \caption{Depth estimation accuracy is sensitive to stereo matching accuracy.}
    \label{fig:sensitive_disp}
  \end{minipage}
\end{figure}

\paragraph{Deconvolution in Stereo DNNs} Stereo matching DNNs implement FE and MO stages as convolution layers. The DR stage fundamentally requires \textit{deconvolution} (a.k.a. transposed convolution) layers~\cite{odena2016deconvolution}. Deconvolution layers generate large activation maps from small input feature maps, essentially up-sampling the input. The up-sampling in DR is critical to compensate the down-sampling in FE and MO that scale down the input images to extract high-level features.

To illustrate the importance of deconvolution in stereo vision,~\Fig{fig:cost_distribution} shows the time distribution of four state-of-the-art stereo matching DNNs across the three stages. The convolution and deconvolution layers combined account for over 99\% of the execution time, from which 38.2\% is attributed to the deconvolution layers.

\paragraph{High Accuracy Stereo Matching} Stereo matching is critical because it generates the disparity, from which depth is estimated (\Equ{eq:triangulation}). Using the industry-standard Bumblebee2 stereo camera~\cite{bumblebee2} as an example (B is \SI{120}{\milli\metre}, \textit{f} is \SI{2.5}{\milli\metre}, and pixel size is \SI{7.4}{\micro\metre}), \Fig{fig:sensitive_disp} shows how the depth estimation error ($y$-axis) varies with the disparity error in pixels ($x$-axis). Different curves correspond to objects at different distances. We find that even two tenths of a pixel error in stereo matching can result in a depth estimation error of 0.5m--5m, which could be catastrophic at the application level.

While existing stereo matching systems achieve high accuracy at the expense of high compute cost, \proj achieves DNN-level accuracy with significantly less compute.

\section{Invariant-based Stereo Matching}
\label{sec:algo}

This section introduces our new \textit{invariant-based stereo matching algorithm} (ISM). 
The key idea of ISM is to exploit the \textit{correspondence invariant} between the stereo images over time. 
After introducing the high-level concept (\Sect{sec:algo:idea}), we then describe the detailed algorithm (\Sect{sec:algo:details}), and discuss important algorithmic design decisions (\Sect{sec:algo:design}). We make the implementation of ISM available at: \url{https://github.com/horizon-research/ism-algorithm}.


\subsection{Overview}
\label{sec:algo:idea}


Stereo matching produces a disparity map~(\Fig{fig:dispmap}), from which depth information is easily obtained through triangulation~(\Fig{fig:triangulation}).
Classic stereo matching algorithms generate the disparity map by matching pixels/features in the left (reference) frame with pixels/features in the right (matching) frame, typically by searching in a finite window. 
However, the accuracy of search-based algorithms is sensitive to the heuristics used in the search, such as feature selection, search window size, matching criterion, etc. 
In contrast, DNN approaches largely avoid heuristics and instead directly learn the matching pairs.
Unfortunately, DNNs come at the cost of a massive increase in compute requirement.

Instead of the binary choice between DNNs and conventional search-based algorithms, we use DNNs to \textit{guide} the search process of classic methods. 
The key observation is that two matched pixels, one from the left image and the other from the right image, correspond to the same point in the physical world. 
While the locations of the two pixels move from frame to frame, they are always projections of the same scene point, and therefore are always a matched pair in any frame. 
In other words, the geometric correspondence relationship between two matched pixels is invariant.

Our new stereo matching algorithm, ISM, exploits this \textit{correspondence invariant} by operating in two modes. 
It obtains stereo correspondences on ``key frames'' through accurate but compute-intensive DNNs. 
The correspondences are then propagated to subsequent non-key frames as good initial guesses to guide the cheaper search-based methods. By combining learnt correspondences with search-based methods that explicitly model the physical world, ISM reduces the total compute cost while retaining DNN-like accuracy.



\subsection{Algorithm}
\label{sec:algo:details}

\begin{figure}[t]
\centering
\includegraphics[width=\columnwidth]{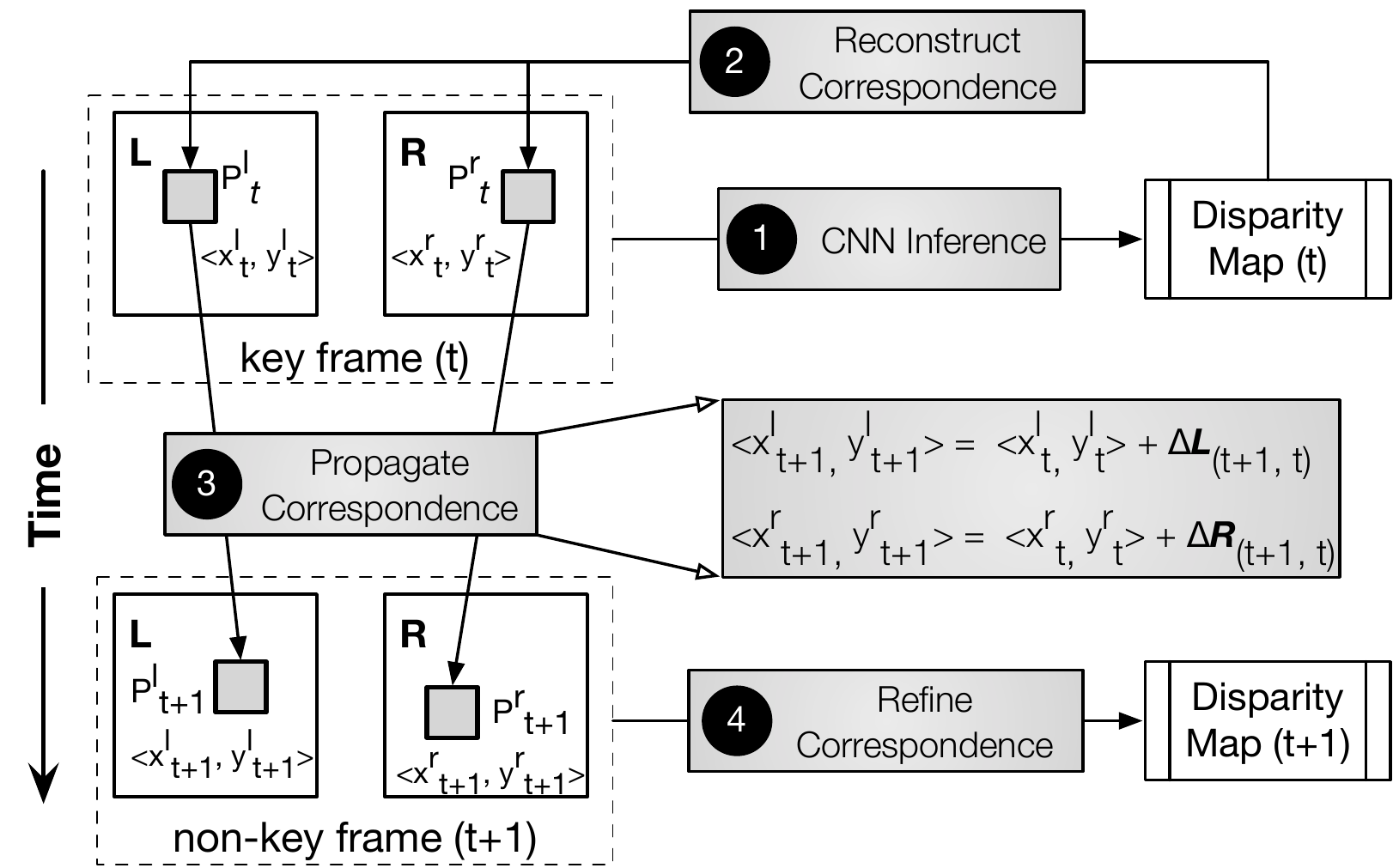}
\caption{The ISM algorithm obtains correspondences in key frames using DNNs, and propagates the correspondences to non-key frames to guide the cheap correspondence search. Time progresses from top to bottom in the figure.}
\label{fig:motion_algorithm}
\end{figure}

We illustrate ISM in \Fig{fig:motion_algorithm}. 
ISM consists of four main components. 
ISM runs DNN inferences (\circled{white}{1}) on key frames to obtain pixel correspondences, which are used to guide feature matching on non-key frames (\circled{white}{2}, \circled{white}{3}, and \circled{white}{4}).



\paragraph{\circled{white}{1} DNN Inference} Assuming the left and right frames at timestep $t$ are regarded as key frames, ISM performs DNN inference to generate a disparity map for the left image, in which each pixel value represents the disparity (i.e., $Z$ in \Fig{fig:triangulation}) of each pixel in the left frame. 
In conventional DNN approaches, this disparity map is used only for triangulation (not shown in the figure) to estimate depth, and is discarded after the depth map is generated.

\paragraph{\circled{white}{2} Reconstruct Correspondences} Instead of discarding the disparity map, ISM uses it to identify the correspondences in the left and right frames. 
As per the definition of disparity (\Equ{eq:dispdef}), every <$x_{t}, y_{t}$> pixel in the disparity map with the value $D_{t}^{\text{<}x, y\text{>}}$ indicates that the <$x_{t}, y_{t}$> pixel in the left frame ($P_{t}^{L}$) and the <$x_{t} + D_{t}^{\text{<}x, y\text{>}}, y_{t}$> pixel in the right frame ($P_{t}^{R}$) form a correspondence pair. 
By iterating through all the pixels in the disparity map, ISM identifies all the correspondence pairs in the left and right frames at timestep $t$.


\paragraph{\circled{white}{3} Propagate Correspondences} A new pair of frames arrives at the next timestep $(t+1)$. 
ISM exploits a well-known observation that pixels in consecutive video frames are highly-correlated in time. 
For instance, $P_{t}^{L}$ has moved to $P_{t+1}^{L}$, and $P_{t}^{R}$ has moved to $P_{t+1}^{R}$. Critically, since $P_{t}^{L}$ and $P_{t}^{R}$ are a correspondence pair projected from a scene point, $P_{t+1}^{L}$ and $P_{t+1}^{R}$ must correspond to the same point, and hence highly likely to also be a correspondence pair at timestep $(t+1)$.

The exact coordinates of $P_{t+1}^{L}$ and $P_{t+1}^{R}$ can be obtained through a \textit{motion estimation} (ME) algorithm. For each pixel in the left (right) frame, the ME algorithm generates a motion vector $\Delta P^{L}_{(t+1, t)}$ ($\Delta P^{R}_{(t+1, t)}$), representing the displacement between the pixel in frame $t$ and frame $(t+1)$. Thus:
$$ P_{t+1}^{L}~=~P_{t}^{L}~+~\Delta P
^{L}_{(t+1, t)} $$
$$ P_{t+1}^{R}~=~P_{t+1}^{R}~+~\Delta P^{R}_{(t+1, t)} $$

\paragraph{\circled{white}{4} Refine Correspondences} Given the correspondence pairs (e.g., $P_{t+1}^{L}$ and $P_{t+1}^{R}$) at timestep $(t+1)$, ISM then calculates the disparity map at $(t+1)$. 
If the motion estimation from $t$ to $(t+1)$ is precise, the propagated correspondence pairs at $(t+1)$ are also precise. 
Accordingly, the disparity map could be simply obtained by calculating the horizontal offsets between all the correspondence pairs. 
For instance, given the correspondence pair $P_{t+1}^{L}$ and $P_{t+1}^{R}$, the disparity at <$x^{l}_{t+1}, y^{l}_{t+1}$> in the disparity map would be $ x^{r}_{t+1} - x^{l}_{t+1}$.

In reality, motion estimation is imperfect due to various visual artifacts such as occlusion and fast motion~\cite{liu1998accuracy}. 
Thus, the correspondences propagated from $t$ are a noisy estimate of the true correspondences at $(t+1)$. 
To further refine the estimate of $(t+1)$ in ISM, we use classic \textit{correspondence search}, and initializes the search window with the propagated correspondences. 
This allows ISM to avoid compute-intensive DNNs on non-key frames without sacrificing accuracy.




\subsection{Algorithmic Design Decisions}
\label{sec:algo:design}

Computing non-key frames requires reconstructing, propagating, and refining correspondences. 
Reconstructing correspondences has little overhead. 
The cost of propagating correspondences is dominated by motion estimation, and the cost of refining correspondences is dominated by the correspondence search. 
Thus, we must carefully choose the motion estimation and correspondence search algorithms such that the compute cost is much lower than DNNs with little accuracy loss. We discuss algorithmic choices below.

\paragraph{Motion Estimation} The literature is rich with motion estimation algorithms, which differ in the coverage and densities of estimated motion. 
The disparity map in stereo matching should ideally be calculated on a per-pixel basis across the frame, so as to enable fine-grained depth estimation. 
This requirement rules out many classic motion estimation algorithms such as block matching (BM)~\cite{bmsurvey}, and sparse optical flow~\cite{lucas1981iterative, horn1981determining}. 
BM estimates motion at the granularity of a block of pixels, and thus does not provide the pixel-level motion that stereo vision requires. Sparse optical flow algorithms such as Lucas-Kanade~\cite{lucas1981iterative} and Horn-Schunck~\cite{horn1981determining} only provide pixel-level motion for feature points such as corners, and do not cover all the frame pixels.

Instead, we use a \textit{dense optical flow} algorithm, specifically the Farneback algorithm~\cite{farneback2003two, farneback2002polynomial}, for motion estimation. Farneback generates per-pixel motion for all the pixels, and is computationally efficient. 99\% of the compute in Farneback is due to three operations: Gaussian blur, ``Compute Flow'', and ``Matrix Update''. Gaussian blur is inherently a convolution operation that convolves a Guassian kernel (2D matrix) with the image. The latter two are point-wise operations that resemble the activation function in DNNs. Thus, motion estimation in the ISM algorithm can be computed using a DNN accelerator to simplify the hardware design.

\paragraph{Correspondence Search} ISM performs correspondence search to refine the initial correspondence estimation propagated through motion. Correspondence search algorithms have been well-studied in the classic computer vision literature~\cite{stereo2002scharstein, stereo2003Brown}, and generally fall into two categories: local methods and global methods. At the cost of higher compute demand, global methods provide higher accuracy by minimizing the pixel motion inconsistencies across the entire image. However, with the initial correspondences propagated through key-frames, we find that local methods suffice.

In particular, we leverage the block matching algorithm~\cite{bmsurvey} for local correspondence search. For each pixel in the left image (e.g., $P_{t+1}^{L}$ in \Fig{fig:motion_algorithm}), ISM uses the block of pixels surrounding it to search in a 1D window in the right image in order to find the closest match. The search window is centered around the initial correspondence estimation (e.g., $P_{t+1}^{R}$ in \Fig{fig:motion_algorithm}). We use the sum of absolute differences (SAD) cost function. The horizontal offset between the two matched blocks is the disparity for $P_{t+1}^{L}$.

Similar to optical flow, the block matching algorithm has a ``convolution-like'' structure~\cite{qadeer2013convolution}; the block in the left image is equivalent to a kernel, and the search window in the right image is equivalent to the input image. 
The only difference is that block matching computes the SAD between the input feature map and the kernel ($\sum^{N}_{i=1} |a_i - b_i|$) as opposed to the dot product in canonical convolution ($\sum^{N}_{i=1} a_i b_i$). 
Thus, the correspondence search can share the same architecture as DNNs and optical flow.

\paragraph{Compute Cost} Due to our algorithmic choices, computation on non-key frames is much cheaper than key-frames. For instance, for a typical qHD frame (960 $\times$ 540), computating a non-key frame requires about 87 million operations while stereo DNN inference (key frame) requires about $10^2\times$--$10^4\times$ more arithmetic operations. 
Thus, ISM leads to significant performance and energy improvements by avoiding DNN inference altogether in non-key frames.

\begin{figure*}[t]
\centering
\includegraphics[width=\textwidth]{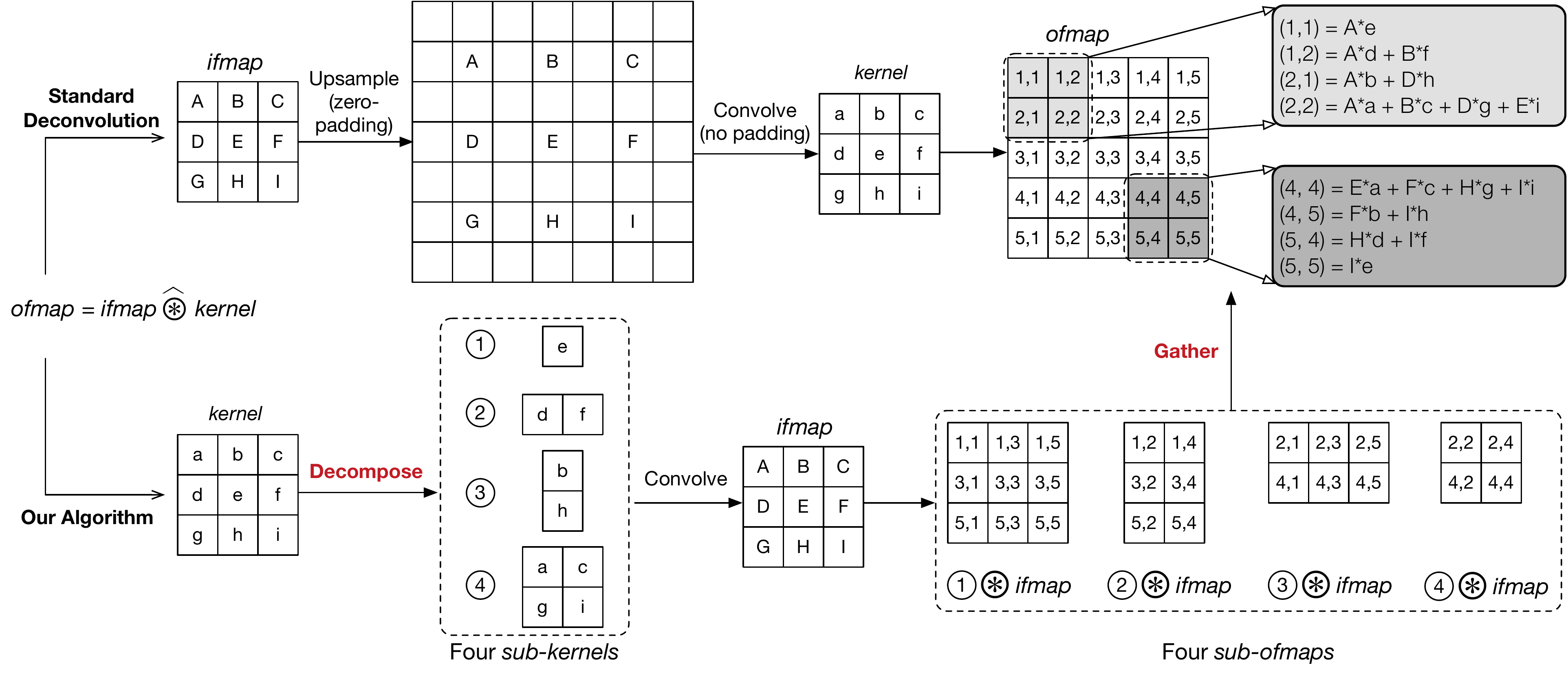}
\caption{Translating deconvolution into multiple convolutions. Standard deconvolution first upsamples the \textit{ifmap} before convolving with the kernel. Note that this example assumes the upsampled \textit{ifmap} is not further padded before the convolution, i.e., a $7 \times 7$ \textit{ifmap} results in a $5 \times 5$ \textit{ofmap}. Our translation algorithm holds regardless of padding.}
\label{fig:split_deconv}
\end{figure*}



\section{Deconvolution Optimizations}
\label{sec:deconv}

While the ISM algorithm removes DNN inference in non-key frames, DNNs remain critical for generating initial key frame correspondences. 
This section describes optimizations for stereo DNNs, in particular the dominant deconvolution layers. 
We propose novel software-only optimizations that mitigate the compute overheads in deconvolution (\Sect{sec:deconv:algo}), while capturing unique data reuse opportunities (\Sect{sec:deconv:reuse}).

We make our optimization framework publicly available at: \url{https://github.com/horizon-research/systolic-array-dataflow-optimizer}. It targets the systolic-array accelerator architecture, supports the deconvolution optimization described here, and applies tiling optimizations to minimize the inference latency and/or DRAM traffic.

\subsection{Deconvolution Transformation}
\label{sec:deconv:algo}

Deconvolution layers on average contribute to 38.2\% (50\% max) of the total MACs in stereo DNNs (\mbox{\Fig{fig:cost_distribution}}). 
Due to the inherent sparsity of deconvolution, a naive mapping to hardware results in over 75\% of redundant computations due to one or more zero operands. 
Deconvolution is also used in Generative Adversarial Networks (GANs), and recent studies have proposed specialized hardware specifically for deconvolution~\cite{song2018GAN, Yazdanbakhsh2018GAN}. 
In contrast to previous studies, we propose a \textit{purely algorithmic transformation} that eliminates inefficiencies due to sparsity. We show that an inherently sparse deconvolution layer can be translated to a series of dense convolutions, which then effectively map on to existing DNN accelerators. 
We next explain the inefficiencies in deconvolution, and then describe our algorithmic transformations.

\paragraph{Standard Deconvolution} The standard process (\Fig{fig:split_deconv}) deconvolves a 3x3 input feature map (\textit{ifmap}) with a 3x3 kernel. 
The \textit{ifmap} is first upsampled with zero padding, before being convolved with the 3x3 kernel to generate an output feature map (\textit{ofmap}). 
Note that the upsampling step essentially performs disparity refinement, which is fundamental to general stereo DNNs, rather than being specific to a particular network~(\Sect{sec:bcg:sv}). The zeros in the upsampled \textit{ifmap} leads to redundant computation and memory traffic.




A key characteristic of deconvolution is that different elements in the \textit{ofmap} are calculated in different ``patterns.''
Consider the first $2\times 2$ outputs in the \textit{ofmap}: (1, 1), (1, 2), (2, 1), and (2, 2). 
Each of the four outputs is generated using a different set of elements from the kernel. 
For instance, (1, 1) requires only $e$ while (1, 2) requires $d$ and $f$. 
Critically, there are only four different patterns, which are repeated across the \textit{ofmap}. 
Pixels (4, 4) and (2, 2) are calculated using the same elements from the \textit{kernel}, as are (1, 1) and (5, 5), (1, 2) and (5, 4), as well as (2, 1) and (4, 5). 
Due to the various patterns needed to generate different output elements, deconvolution is clearly an ``irregular'' operation. 
Prior work~\cite{Yazdanbakhsh2018GAN} exploits the four unique computation patterns by augmenting a conventional DNN accelerator with custom hardware units.

\paragraph{Deconvolution Transformation} In contrast, we find that existing DNN accelerators already provide the necessary architectural substrate to efficiently execute the four different patterns. The key is to recognize that the four computation patterns are essentially four different convolutions, each convolving the original \textit{ifmap} with a distinct kernel that is part of the original kernel. For instance, (2, 2), (2, 4), (4, 2), and (4, 4) are generated by convolving $\begin{bsmallmatrix} a & c \\ g & i  \end{bsmallmatrix}$ with \textit{ifmap}. More generally, the deconvolution in~\Fig{fig:split_deconv} is calculated as:
\begin{equation*}
  \begin{bmatrix}
    a & b & c \\
    d & e & f \\
    g & h & i
  \end{bmatrix}
  \widehat{\circledast}
  ~I = \mathcal{G}(
  \begin{bmatrix}
    e
  \end{bmatrix}
  \circledast
  I, 
  \begin{bmatrix}
    d & f
  \end{bmatrix}
  \circledast
  I, 
  \begin{bmatrix}
    b \\
    h
  \end{bmatrix}
  \circledast
  I, 
  \begin{bmatrix}
    a & c \\
    g & i
  \end{bmatrix}
  \circledast
  I)
\end{equation*}

\noindent where $\widehat{\circledast}$ denotes the deconvolution operation, $\circledast$ denotes the standard convolution operation, $I$ is the \textit{ifmap}, and $\mathcal{G}$ is a gather operation to assemble the \textit{ofmap} from the results of the four convolutions. $\mathcal{G}$ is simply implemented as a set of load operations to the on-chip buffer. Essentially, our algorithm decomposes the original $3 \times 3$ kernel into four sub-kernels, each requiring a smaller dense convolution with the original \textit{ifmap}, which can be executed efficiently on a conventional DNN accelerator.



This transformation generalizes to kernel shapes other than \mbox{$3 \times 3$}. Formally, a 2D kernel $K$ with a dimension $K_H \times K_W$ will be decomposed into four sub-kernels ($S_0, S_1, S_2, S_3$):
\begin{align*}
    S_0^{(i,j)} & = 
                  K^{(2i,2j)}~~~~~~~~~~~~~
    i \in [0, \lceil K_H/2\rceil),
    j \in [0, \lceil K_W/2\rceil) \\
    S_1^{(i,j)} & = 
                  K^{(2i+1,2j)}~~~~~~~~~
    i \in [0, \lfloor K_H/2\rfloor),
    j \in [0, \lceil K_W/2\rceil) \\
    S_2^{(i,j)} & = 
                  K^{(2i,2j+1)}~~~~~~~~~
    i \in [0, \lceil K_H/2\rceil),
    j \in [0, \lfloor K_W/2\rfloor) \\
    S_3^{(i,j)} & = 
                  K^{(2i+1,2j+1)}~~~~~
    i \in [0, \lfloor K_H/2\rfloor),
    j \in [0, \lfloor K_W/2\rfloor) 
\end{align*}


\noindent where $ S_{*}^{(i,j)} $ is the element $(i, j)$ in a particular sub-kernel, and $ K^{(*,*)} $ is an element in the original kernel $K$. For instance, $ S_0^{(i,j)} = K^{(2i,2j)} $ means that element $(i, j)$ in the first sub-kernel comes from element $(2i, 2j)$ in the original kernel. The boundary condition of each case denotes the dimension of the corresponding sub-kernel (notice the different floor and ceiling functions in each). Hence, decomposing a $3 \times 3$ kernel results in four sub-kernels of shapes $2 \times 2$, $1 \times 2$, $2 \times 1$, and $1 \times 1$, confirming the specific example above. The general formulation of the deconvolution transformation with an arbitrary N-dimensional kernel is described in \Apx{sec:appendix:decompose}.



\subsection{Exploiting Inter-Layer Activation Reuse}
\label{sec:deconv:reuse}

A beneficial trait of our transformation is that each sub-convolution reads the same \textit{ifmap}, which in modern DNNs does not fit in on-chip buffers and must spill to main memory. In contrast, our transformation can uniquely exploit \textit{inter-layer activation reuse} because each sub-convolution layer shares the same \textit{ifmap}. The challenge is to systematically maximize the reuse exploited across the entire network while minimizing the inference latency.


We primarily consider \textit{loop tiling}, which is known to be critical to exploiting data reuse in DNNs~\cite{Mullapudi2016AutoHalide, yang2018dnn}. Prior work in DNN tiling predominately searches for the tiling strategy in a brute-force manner~\cite{Ma2017dnnFPGA, Hegde2018morph}. However, brute-force search does not scale to stereo DNNs for two reasons. First, our translation scheme significantly increases the number of layers, each of which must be individually searched. For instance in the example of \Fig{fig:split_deconv}, the number of layers quadruples; a 3D kernel could increase layers by $8\times$. Second, exploiting the inter-layer \textit{ifmap} reuse adds another scheduling dimension, further increasing the search space.


Instead of a search, we formulate the reuse optimization as a constrained optimization problem, minimizing layer latency while satisfying hardware resource constraints. Our optimization can be efficiently solved using a greedy algorithm.


\paragraph{Architectural Assumptions} We first describe the underlying architecture that the optimization formulation assumes. 
Overall, we make standard assumptions that generally hold across the vast majority of current DNN accelerators. \Sect{sec:sys:hw} describes the hardware architecture in detail.

We assume a systolic array accelerator. Each Processing Element (PE) performs one MAC operation per cycle~\cite{jouppi2017datacenter, samajdar2018scale}. Systolic arrays use a very efficient neighbor-to-neighbor communication mechanism, particularly well suited to convolution. Alternatively, our formulation could also be extended to support spatial arrays~\cite{Eyeriss2016}, which offer more flexible control at higher hardware cost.

We assume that the accelerator has a unified on-chip buffer (scratchpad) for the \textit{ifmap}, kernels, and \textit{ofmap}. This buffer is generally too small to hold all the data for a whole layer. Therefore, the \textit{ofmap} is computed in multiple rounds. Only part of the \textit{ifmap} and the kernels are stored in the buffer each round. 
The optimal scheduling of partial \textit{ifmap} and kernels in the buffer for each round is critical to maximizing reuse.

The buffer is evenly split into working and filling sections for double-buffering. 
While the PE array is computing the current round using data in the working buffer, the data for the next round is pre-fetched to the filling buffer. The next round starts only when the filling buffer is full. This design choice guarantees that any data access by the PEs will hit in the buffer without stalling the PE array.

\paragraph{Optimization Formulation} We follow a layer-wise execution model, in which a layer only starts after the previous layer finishes. Therefore, minimizing the total latency is equivalent to minimizing the latency of each individual layer. We describe how the latency of a deconvolution layer is formulated and optimized. Our formulation can be easily extended to support a convolution layer, which can be regarded as a special case of deconvolution without \texttt{ILAR}.

The optimization objective is to minimize the deconvolution layer's latency given hardware resource constraints. Note that since a deconvolution is translated to a set of convolutions, it is the cumulative latency of these sub-convolutions that is of interest. The optimization problem is formulated as follows:
\begin{align}
    \min\ &\ L(\Theta, \phi) \\
    s.t.\ &\ R(\Theta) \leq R^*
\end{align}

\begin{figure}[t]
\centering
\includegraphics[width=\columnwidth]{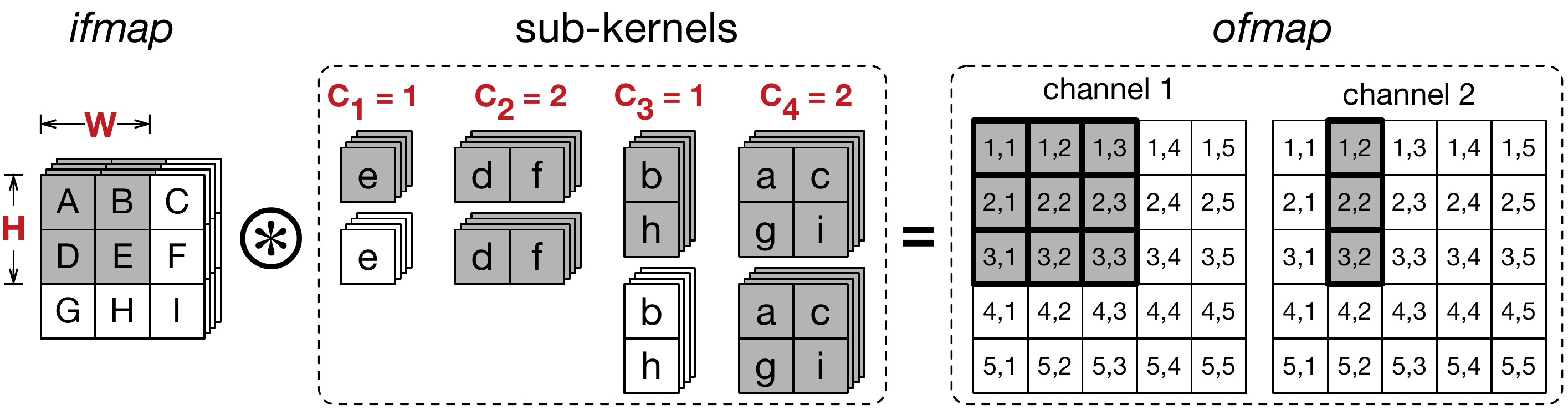}
\caption{Tiling in a translated deconvolution with a $3 \times 3$ kernel split into four sub-kernels. With a tiling strategy $W=2, H=2, C_1=1, C_2=2, C_3=1, C_4=1$, only the shaded elements are loaded into the buffer. The \textit{ofmap} elements generated in this round (shaded) are also stored in the buffer.}
\label{fig:tiling}
\end{figure}


\noindent where $\Theta$ denotes a particular hardware configuration, and $R(\cdot)$ is the configuration's hardware resources, which must not exceed the specified resource budget $R^*$. We consider three main types of hardware resources: 1) PE array size, 2) on-chip buffer size, and 3) off-chip memory bandwidth. 

Latency $L(\cdot)$ is affected by both the hardware configuration ($\Theta$) and the tiling schedule ($\phi$). The optimal tiling is determined by the following variables: 1) the dimension of the \textit{ifmap} tile to be loaded into the buffer ($W$ and $H$), and 2) the number of filters in each \textbf{sub-kernel $k$} to be loaded into the buffer ($C_k$). Critically, $C_k$ can be different for each sub-kernel. \Fig{fig:tiling} illustrates these optimization variables, with an example where part of the \textit{ifmap} is convolved with certain filters of the four sub-kernels to generate a partial \textit{ofmap}. 
The vector $\overrightarrow{C}$ denotes the collection of all $C_k$.

With double buffering, a layer $L$'s latency is the cumulative latency across all $N$ rounds. The latency of each round ($l^i$) is determined by the maximum value between the memory access time ($l^{i}_{m}$) and the compute time ($l^{i}_{c}$) of the round:
\begin{align}
L(\Theta, \phi) = \sum^{N}_{i=1} l^{i}(\Theta, \phi),~~~~~~l^{i}(\Theta, \phi) = \max(l^{i}_{c}, l^{i}_{m})
\end{align}



With double-buffering, $l^{i}_{c}$ is determined by two sets of parameters: 1) $W^{i}$, $H^{i}$, and $\overrightarrow{C^{i}}$, which decide the total compute demand, and 2) the PE array size, $A^*$, which decides the compute capability. $l^{i}_{c}$ is the cumulative latency of processing each individual sub-kernel:
\begin{align}
    l^i_{c} & = \sum_{k=1}^{|\overrightarrow{C^i}|} \ceil{\frac{W^i_k \times H^i_k \times I \times C^i_k \times H^i \times W^i}{ A^*}}
\end{align}
\noindent where $|\overrightarrow{C^i}|$ denotes the total number of sub-kernels in round $i$, $W^i$ and $H^i$ are the dimensions of the \textit{ifmap} tile loaded into the buffer in round $i$, $W^i_k$ and $H^i_k$ are the dimensions of sub-kernel $k$ in round $i$\footnote{The sub-kernels' dimensions do not change across rounds. Given a $k$, $W^i_k$ and $H^i_k$ are constants for any $i$. For the consistency of the notations, we still use $W^i_k$ and $H^i_k$.}, $C^i_k$ denotes the number of filters in sub-kernel $k$ loaded into the buffer in round $i$, and $I$ is the number of input channels. The ceil operator indicates that the next sub-kernel can not start until the previous sub-kernel is finished even if the PE array is under-utilized. This is because only one sub-kernel can be calculated on the systolic array at a time as sub-kernels vary in their shapes.

The memory access time, $l^i_{m}$, is determined by the available memory bandwidth, $B^*$, and the amount of data that needs to be transferred to/from DRAM each round, which in turn depends on the reuse order: whether the \textit{ifmap} tile or the sub-kernels remain in the buffer across consecutive rounds. A binary variable $\beta$ denotes this reuse order, and $l^i_{m}$ becomes:
\begin{align}
    l^i_{m} & = \beta \times l^i_{m:W} + (1-\beta) \times l^i_{m:In},\ \beta \in \{0,1\}
\end{align}
where $ l^i_{m:In} $ is the memory access latency if the \textit{ifmap} remains in the buffer, and $l^i_{m:W}$ denotes the memory latency if the sub-kernels remain in the buffer. Specifically:
\begin{align}
    l^i_{m:W} & = \ (\Delta IF^i + \sum_{k = 1}^{|\overrightarrow{C^i}|} \Delta OF^i_{k}) \times\frac{1}{B^*} \\
    l^i_{m:In} & = \ \sum_{k = 1}^{|\overrightarrow{C^i}|} (\Delta W^i_{k} + \Delta OF^i_{k}) \times \frac{1}{B^*}
\end{align}
\noindent where the terms with prefix $ \Delta $ denote the amount of data that needs to be loaded from DRAM. Depending on the reuse order, either the \textit{ifmap} elements ($\Delta IF^i$), or the sub-kernels ($\Delta W^i_k$) are loaded. The newly computed \textit{ofmap} elements ($\Delta OF^i_k$) are always stored back to DRAM. Note that $\Delta W^i_k$, $\Delta IF^i$, and $\Delta OF^i_k$ are all deterministic functions of $W^i_k$, $H^i_k$, $W^i$, $H^i$, and $|\overrightarrow{C^i}|$. We omit them here for brevity, and describe their exact expressions in~\Apx{sec:appendix:exp}.

The on-chip buffer capacity ($Buf^*$) imposes the constraint:
\begin{equation}
   \Delta IF^i + \sum_{k = 0}^{|\overrightarrow{C^i}|} (\Delta OF^i_k + \Delta W^i_k) \leq Buf^*
\end{equation}

Finally, $C^i_k$ and $N$ must satisfy:
\begin{align}
\forall k \in \{1, 2, ..., |\overrightarrow{C}|\}, \mathbb{C} = \sum^{N}_{i=1} C^i_k
\end{align}
\noindent where $\mathbb{C}$ denotes the number of output channels of a layer, which is a constant invariant to $k$ and $i$.

Overall, this formulation minimizes the latency $L$ with respect to $W^i$, $H^i$, and $C^i_k$ ($i \in \{1,2,...,N\}$, $k \in \{1, 2, ..., |\overrightarrow{C}|\}$), under the hardware resource constraints $A^*$, $B^*$, and $Buf^*$.

\paragraph{Efficient Solver} The above constrained-optimization problem has non-convex objective and constraints, and thus has no closed-form solutions. To derive a solution efficiently, we convert this problem to a Knapsack-like structure, where each filter in each sub-kernel is an \textit{item}, the size of each filter is the \textit{weight}, and the number of MAC operations associated with each filter is the \textit{value}.

To solve the Knapsack problem, we use a simple greedy heuristic that prioritizes filters from large sub-kernels with standard dynamic programming. In contrast to the classic 0/1 Knapsack problem, our problem formulation requires us to consume all the items, since all the filters in each sub-kernel are required to finish a convolution. We therefore iteratively apply the greedy solver until all the items are used. The solver is executed offline, and finishes within one second on an Intel Core i5-7500 CPU.



\begin{figure}[t]
\centering
\includegraphics[width=1\columnwidth]{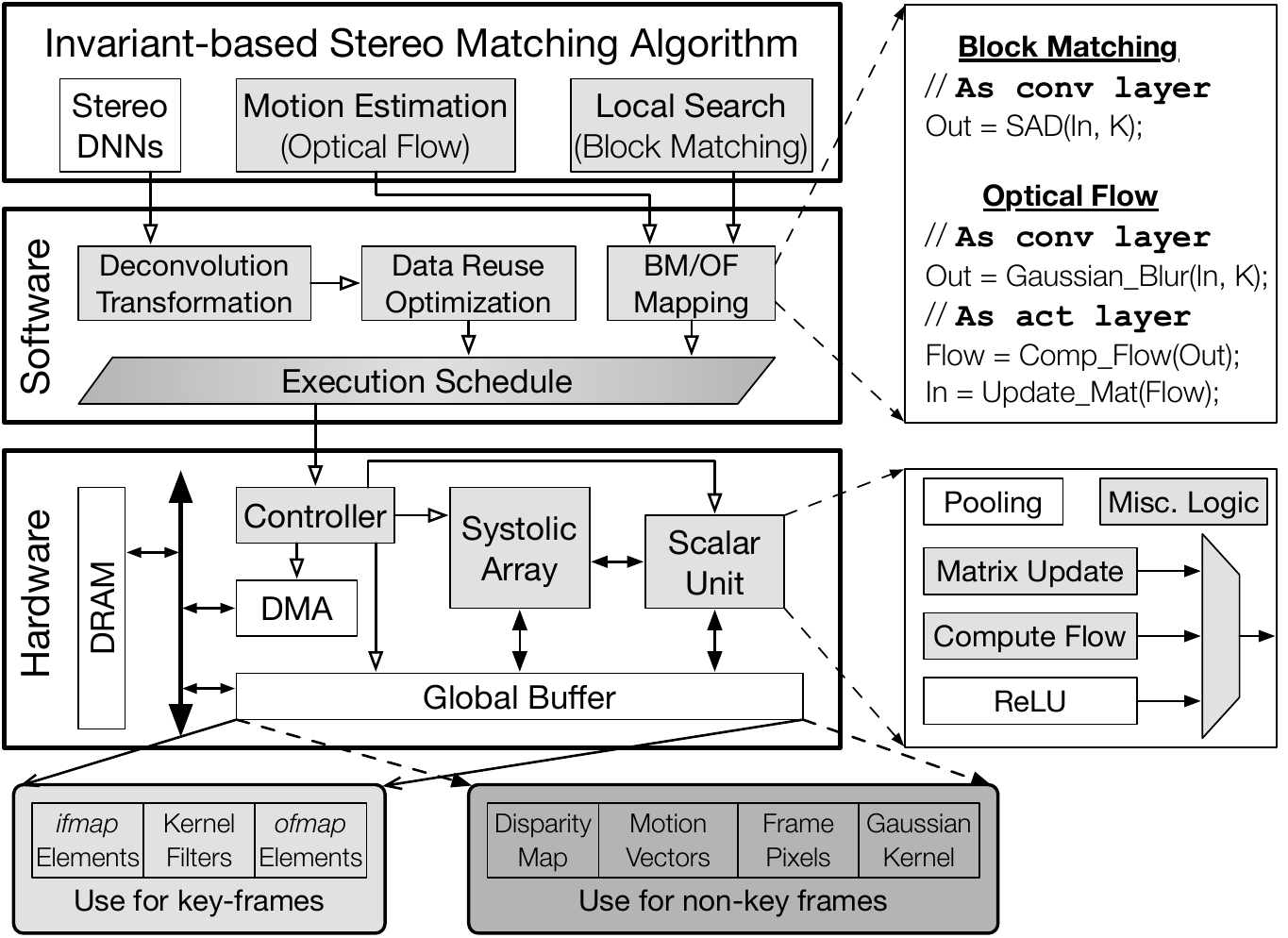}
\caption{The \proj overview with augmentations shaded.}
\label{fig:system}
\end{figure}

\section{The ASV System}
\label{sec:sys}

Building on top of the ISM algorithm and the deconvolution optimizations, this section presents the software and hardware system of \proj.~\Fig{fig:system} gives a high-level system overview of \proj. We first present the software system (\Sect{sec:sys:sw}), and then discuss the architecture design decisions (\Sect{sec:sys:hw}).

\subsection{Software System}
\label{sec:sys:sw}


The goal of the software system in \proj is to map the ISM algorithm to the underlying hardware. 
The static mapping is done offline. 
There are three components in the ISM algorithm to map: stereo matching DNN, motion estimation, and local correspondence search. We rely on the user to supply a particular stereo DNN depending on their accuracy needs. Motion estimation and correspondence search are implemented using optical flow (OF) and block matching (BM), respectively, as described in \Sect{sec:algo:design}. We now describe how each component is processed by the software.

\paragraph{Mapping Stereo Matching DNN} For the deconvolution layer, the \proj software performs the deconvolution transformation, as well as the data reuse optimization. For convolution layers, while the deconvolution transformation does not apply, we apply the data reuse optimization \textit{without} ILAR. In the end, we obtain a transformed stereo DNN along with an execution schedule, which are both consumed by the hardware at runtime. The schedule includes the tiling strategy and buffer partitioning strategy for each layer.

\paragraph{Mapping OF/BM} The software maps the OF and BM algorithms in ISM to a set of convolution and/or activation operations that are directly interfaced with conventional DNN accelerators. The software translates the BM operation to a convolution layer (\Sect{sec:algo:design}), but calculating SAD instead of dot product at each window.

The OF computations include Gaussian blur, ``Compute Flow'' and ``Matrix Update'' operations (\mbox{\Sect{sec:algo:design}}), shown in the top-right box in~\mbox{\Fig{fig:system}}. Gaussian blur is naturally expressed as a convolution layer with one output channel. ``Compute Flow'' and ''Matrix Update'' are point-wise operations expressed as special activation functions.



\subsection{Hardware Architecture}
\label{sec:sys:hw}

Leveraging the software pass, the hardware requires only minimal, structured augmentations on top of a conventional DNN accelerator. We start from a baseline DNN accelerator and describe how the compute, memory, and control logic is augmented with \proj-specific architectural extensions.

\paragraph{Compute} Our baseline DNN accelerator consists of a TPU-like systolic PE array for convolution and a scalar unit for non-convolution operations, e.g., activation~\mbox{\cite{jouppi2017datacenter}}. Each PE consists of two 16-bit input registers, a 16-bit fixed-point MAC unit with a 32-bit accumulator register, and simple trivial control logic. This is identical to the PE in the TPU~\mbox{\cite{jouppi2017datacenter}}.

We use the systolic array as the baseline due to its efficiency in handling convolutions. However, our software optimizations do not depend on a particular baseline DNN architecture. Alternatives such as more flexible spatial architectures~\mbox{\cite{Eyeriss2016,chen2014diannao}} are also suitable, albeit requiring different constrained-optimization formulations to those presented in~\mbox{\Sect{sec:deconv:reuse}}. We will later demonstrate the effectiveness of our deconvolution optimizations on Eyeriss~\mbox{\cite{Eyeriss2016}}.

\mbox{\proj} augments both the systolic array and the scalar unit in the baseline architecture to support the ISM algorithm. First, each PE is extended with the capability to accumulate absolute differences (i.e., $a \leftarrow a +|b - c|$) in addition to MAC in order to support BM. Second, we extend the scalar unit to support two additional point-wise operations: ``Compute Flow'' and ``Matrix Update''; both are required by OF (as illustrated in the bottom-right box in~\mbox{\Fig{fig:system}}).

Finally, the hardware includes a very small amount of additional logic to support the remaining operations in the ISM algorithm that are inefficient to map to either the systolic array or the point-wise scalar unit. 
These operations are comparisons and control-flow, and are orders of magnitude less costly in area and power compared to the systolic array and the scalar unit. 
For instance, BM requires comparing the SAD values across different matched blocks, and OF requires checking the value boundaries during "Matrix Update".

\paragraph{Memory} \proj uses the familiar three-level memory hierarchy~\cite{li2019mlmem}. Each PE has a small register file to exploit intra/inter-PE data reuse. A DMA engine coordinates data transfer between the on-chip global buffer and off-chip memory. The global buffer is temporally-shared between key frames and non-key frames. When processing key frames, the global buffer holds the ifmap, kernels, and ofmaps. The exact buffer partitioning is dictated by the \proj software.

When processing non-key frames, the global buffer holds four pieces of data: the pixels of the current and key frames, the Gaussian kernel, the motion vectors, and the disparity maps. The frame pixels dominate the storage requirement, but could be tiled because they are used in Gaussian blur and BM, both of which are convolution operations. The rest of the data cannot be tiled, and thus imposes a minimum buffer size. Assuming qHD resolution (960 $\times$ 540), we enforce a minimum buffer size of about 512 KB.


\paragraph{Control} A micro-sequencer is used to coordinate the computation and memory accesses. In \proj, the sequencer also chooses key frames. Although complex adaptive schemes are feasible~\cite{zhu2018euphrates, buckler2018eva2}, we found that a simple strategy to statically set the key-frame window suffices (\Sect{sec:eval:ac}).



\section{Evaluation Methodology}
\label{sec:exp}

This section introduces the basic hardware and software setup~(\Sect{sec:exp:setup}), and outlines the evaluation plan~(\Sect{sec:exp:plan}).

\subsection{Basic Setup}
\label{sec:exp:setup}

\paragraph{Hardware Implementation} We develop validated RTL implementations for the \proj hardware. The hardware is based on a systolic array architecture, consisting of $24\times24$ PEs clocked at 1~GHz. Each PE is capable of performing both the MAC and absolute difference operations. The hardware also has a scalar unit clocked at 250~MHz, which consists of 8 parallel lanes, each capable of performing the ReLU activation function as well as the point-wise matrix update and compute flow operations required by OF. The on-chip buffer (SRAM) is 1.5~MB in size and is banked at a 128~KB granularity. While we primarily evaluate \proj using this configuration, we will later show sensitivity of \proj performance to different hardware resource configurations.


The RTL is implemented using Synposys synthesis and Cadence layout tools in TSMC 16nm FinFET technology, with SRAMs generated by an ARM compiler. Power is simulated using Synopsys PrimeTimePX, with full annotated switching activity. The off-chip DRAM is modeled after four Micron 16 Gb LPDDR3-1600 channels~\cite{micronlpddr3}. Overall, the accelerator layout has a total area of \SI{3.0}{\mm\squared}, and produces a raw throughput of 1.152 Tera operations per second.

\paragraph{Stereo DNNs} The ISM algorithm can use an arbitrary stereo DNN. We evaluate four state-of-the-art DNNs:  \textsc{FlowNetC}~\cite{FlownetC2015}, \textsc{DispNet}~\cite{DispNet2015}, \textsc{GC-Net}~\cite{NvStereo2018}, and  \textsc{PSMNet}~\cite{PSMNet2018}, with varying accuracy--performance trade-offs (\Fig{fig:trade-off}). 

\paragraph{Dataset} We evaluate \proj on two widely-used datasets: SceneFlow~\cite{DispNet2015} and KITTI~\cite{menze2015kitti}. SceneFlow contains 26 pairs of synthetic stereo videos to mimic various scenarios with different depth ranges. KITTI contains 200 pairs of stereo frames captured from real street views that cover varying driving scenarios and conditions. 

We use the standard ``three-pixel-error'' accuracy metric~\cite{menze2015kitti, kitti_benchmark}, which considers a pixel's depth to be correct if its disparity error is less than 3 pixels compared to ground truth. We then report the percentage of correct pixels, following the convention in the vision and robotics literature~\cite{PSMNet2018, NvStereo2018, DispNet2015, GCNet2017}.



\begin{figure}[t]
\centering
\includegraphics[width=\columnwidth]{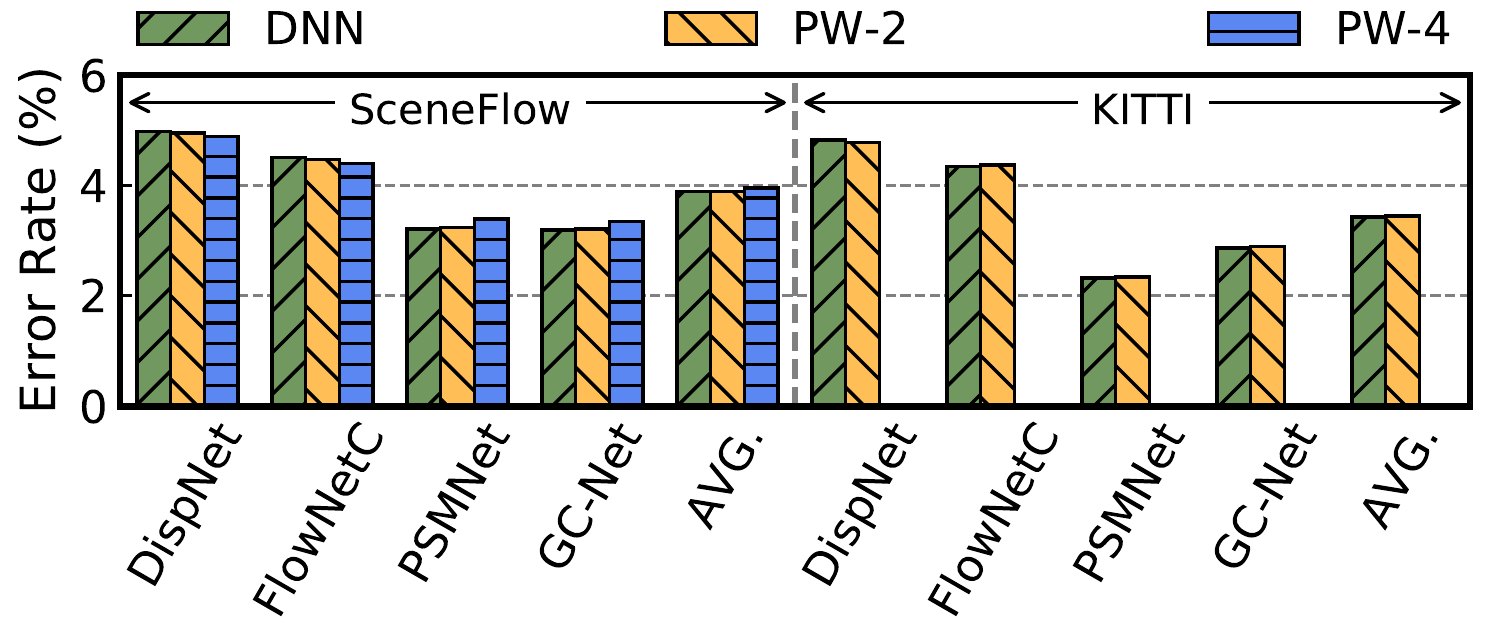}
\caption{Error rate comparison between the ISM algorithm in \proj and the DNN baselines.}
\label{fig:ISM_result}
\end{figure}

\subsection{Evaluation Plan}
\label{sec:exp:plan}

Our goal is to demonstrate the effectiveness of \proj over generic CNN accelerators that are not optimized for stereo vision workloads. We separate the efficiency gains of the new ISM algorithm from that of the deconvolution optimizations.

\paragraph{Baselines} Our baseline is a generic systolic array CNN accelerator, which executes stereo DNNs without any \proj optimizations. 
Today's CNN accelerators mostly statically partition the on-chip buffer across \textit{ifmap}, weights, and \textit{ofmap}. To obtain a strong baseline, we determine the partitioning strategy by exhaustively searching all the partitions offline and use the one that achieves the lowest latency for the entire DNN. Note that the same partition is used for all the layers whereas our data reuse optimization generates different partitions for different layers.

We also compare against Eyeriss~\cite{Eyeriss2016}, a DNN accelerator based on a more flexible spatial architecture. Eyeriss performance and energy are obtained using the public simulator~\cite{gao2017tetris, eyerisssim}. For a fair comparison, we configure Eyeriss to have the same PE counts, on-chip memory capacity, and memory bandwidth as \proj. Finally, to establish a baseline, we also show the results of the Pascal mobile GPU found in the 16~nm Nvidia Parker SoC hosted on the Jetson TX2 development board~\cite{tx2dev}. We use the built-in power sensing circuity to obtain the energy consumption.

\paragraph{\proj Variants} We present an ablation study on \proj to separate the gains from different optimizations:

\begin{itemize}[topsep=0pt]
  \item \mode{ISM}: ISM algorithm without deconv. optimizations.
  \item \mode{DCO}: Deconv. optimizations without ISM algorithm.
  \item \mode{ISM+DCO}: Both ISM and deconv. optimizations.
\end{itemize}

\section{Evaluation}
\label{sec:eval}


We first show that \proj adds negligible overhead to the baseline DNN accelerator (\Sect{sec:eval:oh}) and introduces negligible accuracy loss (\Sect{sec:eval:ac}). We then show the performance and energy improvements of \proj (\Sect{sec:eval:df}), which are robust against the underlying hardware configuration (\Sect{sec:eval:sensitivity}). \proj also out-performs Eyeriss and GPUs (\Sect{sec:eval:eyeriss}). Finally, we demonstrate the general applicability of our deconvolution optimizations by showing that they even improve runtime of GANs without hardware modifications (\Sect{sec:eval:gannx}).

\subsection{Hardware Overhead}
\label{sec:eval:oh}

Owing to the software transformations, \proj only minimally augments existing DNN accelerators. Relative to the baseline accelerator, \proj extends each PE to support accumulating absolute difference. This adds 6.3\% area (\SI{15.3}{\micro\metre\squared}) and 2.3\% power (\SI{0.02}{\milli\watt}) overhead \textit{per PE}. \proj also extends the scalar unit to support new point-wise operations, with an area and power overhead of \SI{2}{\mm\squared} and \SI{2.2}{\milli\watt}, respectively. The overall area and power overhead introduced by \proj are both below 0.5\%.



\subsection{Accuracy Results}
\label{sec:eval:ac}

\proj matches or even outperforms DNN accuracy.~\Fig{fig:ISM_result} shows the accuracy of applying the ISM algorithm to stereo matching DNNs. We use \textit{Propagation Window} (PW) to denote how far in time the correspondence invariant is propagated, which in turn decides how often key frames are selected. With PW-2, every other frame is selected as a key frame, and for PW-4, every fourth frame is a key frame. Note that the KITTI dataset contains at most two consecutive frames, and thus we evaluate only PW-2.

On both datasets, PW-2 retains the same accuracy as the stereo DNNs. On SceneFlow, PW-4 results in only 0.02\% accuracy loss. In some cases, ISM combined with the DNNs can outperform the DNNs alone. For instance, applying the ISM algorithm with \textsc{FlowNetC} reduces error by 0.11\% at PW-4. Overall, our experiments shows that by leveraging the correspondence invariant over time, ISM is able to preserve the DNN-like accuracy with cheap, classic stereo matching algorithms. We will now show that \proj achieves high accuracy while greatly improving the performance and energy-efficiency of stereo vision.

\subsection{Speedup and Energy Reduction}
\label{sec:eval:df}

\begin{figure}[t]
\centering
\includegraphics[width=\columnwidth]{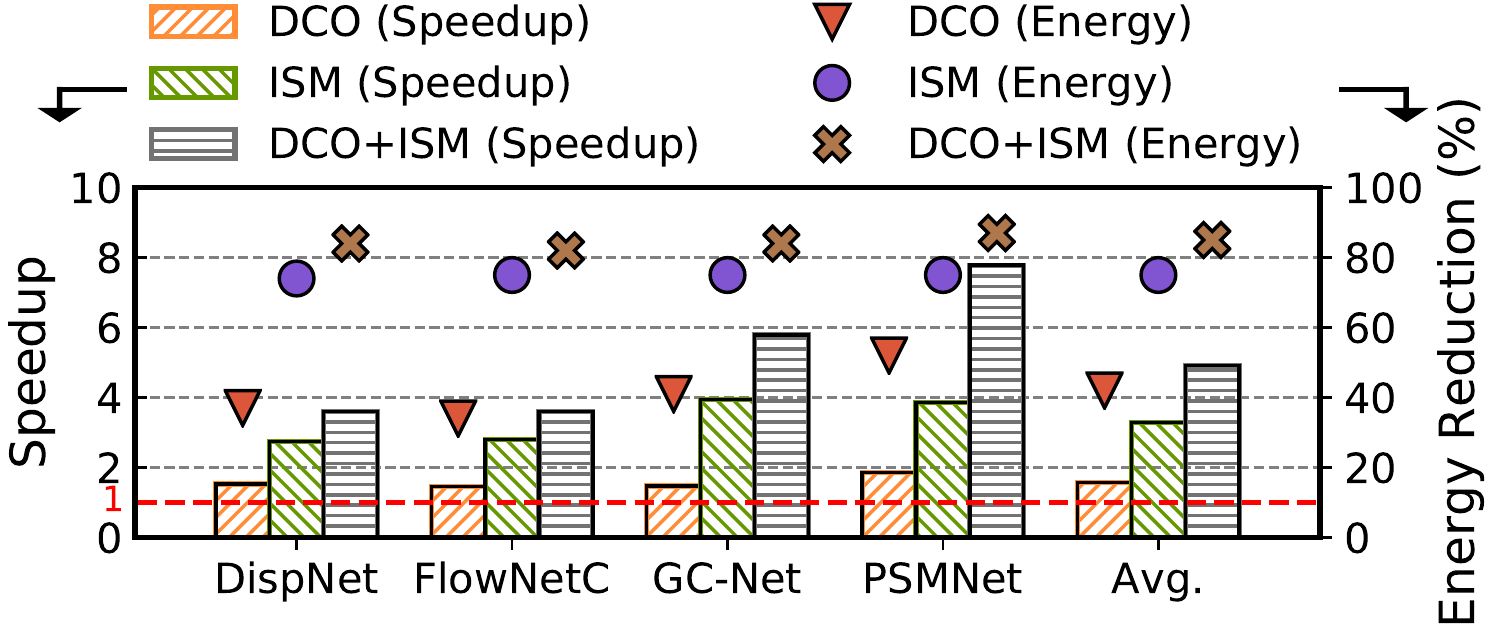}
\caption{Speedup and energy reduction of the three \proj variants over the baseline.}
\label{fig:asv_result}
\end{figure}

\proj significantly improves performance and energy consumption of stereo vision. To understand the contributions of the ISM algorithm and the deconvolution optimizations,~\Fig{fig:asv_result} shows the speedup and energy reduction of the three ASV variants (\Sect{sec:exp:plan}) over the baseline, when applied to different stereo DNNs. We choose PW-4 for the ISM algorithm. On average, combining ISM and deconvolution optimizations (DCO) \proj achieves 4.9$\times$ speedup and 85\% energy reduction. Specifically, ISM achieves, on average, 3.3$\times$ speedup and 75\% energy reduction, while DCO achieves 57\% performance improvement and 38\% energy reduction. ISM contributes more than DCO because ISM avoids DNNs in non-key frames altogether by using the much cheaper BM and OF algorithms (\Sect{sec:algo:design}).

Next, we dissect different optimization components within DCO to further understand the effect of each optimization.

\begin{figure}[t]
\centering
\subfloat[Speedup and energy reduction on deconvolution layers only.]{
	\label{fig:dnns_deconv}
	\includegraphics[width=\columnwidth]{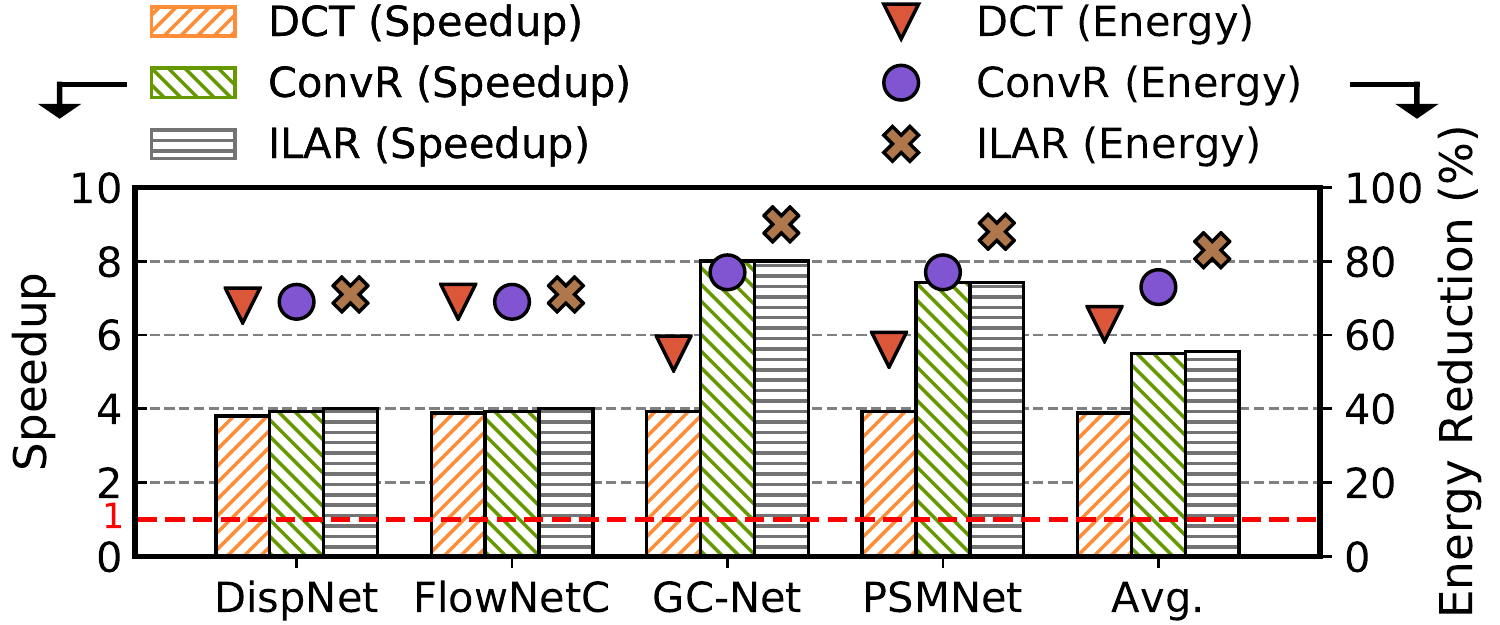} }
\\
\subfloat[Speedup and energy reduction on the entire network.]{
	\label{fig:dnns_overall}
	\includegraphics[width=\columnwidth]{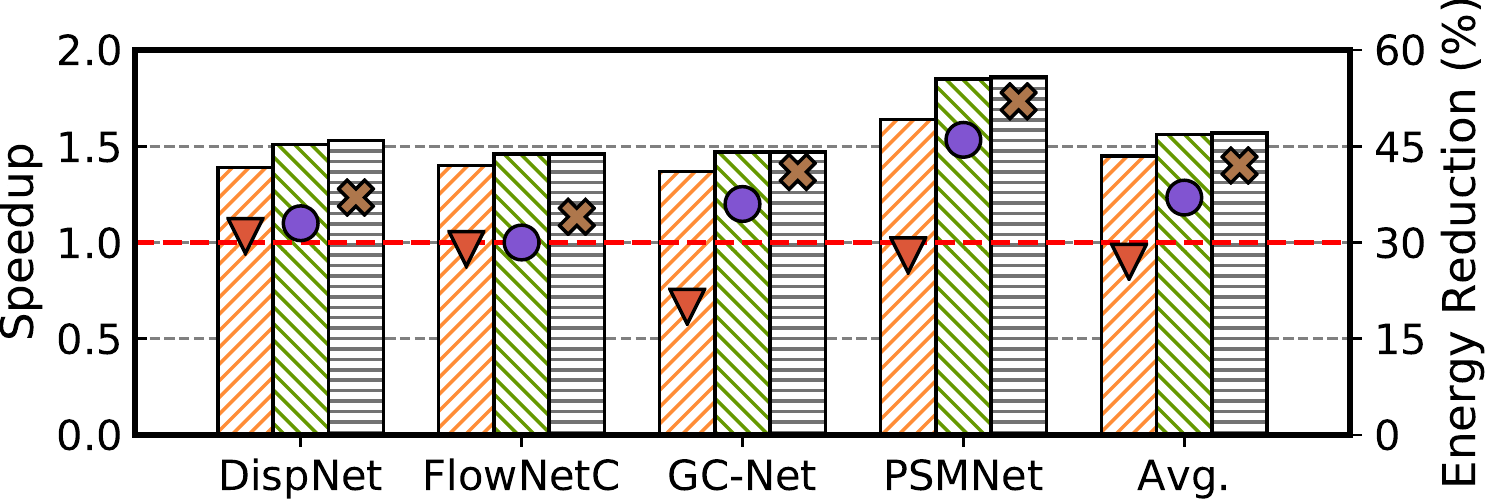} } 
\caption{The speedup and energy reduction of various deconvolution optimizations. Higher is better.}
\label{fig:dnns_result}
\end{figure}

\paragraph{Deconvolution Optimizations} Deconvolution optimizations consist of two components: the deconvolution to convolution transformation (\texttt{DCT} - \Sect{sec:deconv:algo}) and the data-reuse optimization (\Sect{sec:deconv:reuse}). In particular, our data-reuse formulation unifies the exploitation of two kinds of reuse: the conventional data reuse in convolution layers and the inter-layer activation reuse in deconvolutions that is uniquely exposed by \texttt{DCT}. To clearly tease apart our contributions, we show the results of both the conventional reuse optimization (\texttt{ConvR}), which is obtained by applying our reuse optimizer (\Sect{sec:deconv:reuse}) \textit{without} exploiting inter-layer activation reuse, and the additional effect of exploiting inter-layer activation reuse (\texttt{ILAR}).

\Fig{fig:dnns_result} shows the speedup and energy reduction of \texttt{DCT}, \texttt{ConvR}, and \texttt{ILAR}.~\Fig{fig:dnns_deconv} shows the improvements of deconvolution layers only, and~\Fig{fig:dnns_overall} shows the improvements of the entire network. The majority of speedup is from deconvolution transformation, which yields an average 3.9$\times$ speedup on deconvolution layers alone and 1.4$\times$ speedup on the entire network. On top of \texttt{DCT}, \texttt{ConvR} and \texttt{ILAR} further increase speedup to 5.6$\times$ and 1.6$\times$ on deconvolution layers alone and the entire networks, respectively.


Across different stereo DNNs, we find that 3D DNNs (\textsc{GC-Net} and \textsc{PSMNet}) have a speedup of 7.7$\times$ on deconvolution layers, higher than the 3.9$\times$ speedup of 2D DNNs (\textsc{DispNet} and \textsc{FlowNetC}). The reason is twofold. First, 3D DNNs have the higher percentage of zero-padding than 2D DNNs (8$\times$ vs. 4$\times$), which are effectively eliminated by our deconvolution transformation. Second, after the deconvolution transformation the 3D DNNs have many small kernels (e.g., 1$\times$1$\times$1), which leads to low data-reuse. Thus, reuse optimizations become more critical to these networks. In contrast, most 2D stereo DNNs inherently have better data reuse with larger kernels (e.g., 5$\times$5). We also observe that \texttt{ConvR} and \texttt{ILAR} have similar performance. This is because both optimize the data reuse to the extent that the layer processing becomes limited by the PE size.


While \texttt{ILAR} is similar in speedup compared to \texttt{ConvR}, \texttt{ILAR} is much more effective in reducing energy than \texttt{ConvR}. To demonstrate this,~\Fig{fig:dnns_result} overlays the energy reductions of different DCO variants on the right $y$-axis. DCO achieves 83\% energy reduction on deconvolution alone and 38\% on the entire network. Specifically, \texttt{DCT} reduces the deconvolution energy by 62\%; \texttt{ConvR} and \texttt{ILAR} further improve the energy reduction to 73\% and 83\%, respectively.

The energy saving of \texttt{DCT} comes from eliminating redundant movement of padded zeros in the upsampled \textit{ifmap}. \texttt{ILAR} achieves additional energy reduction over \texttt{ConvR} by exploiting inter-layer activation reuse, a unique reuse behavior in our transformed deconvolution layers. 3D DNNs benefit much more from \texttt{ILAR} than 2D DNNs, as is evident by examining the additional energy reductions of \texttt{ILAR} over \texttt{ConvR} across networks. This is because 3D stereo DNNs have low \textit{ifmap} data-reuse; \texttt{ILAR} uniquely exploits inter-layer \textit{ifmap} reuse, and thus reduces more memory traffics.

\subsection{Sensitivity Analysis}
\label{sec:eval:sensitivity}

\begin{figure}[t]
  \centering
  \captionsetup[subfigure]{width=0.5\columnwidth}
  \subfloat[\small{Speedup.}]
  {
  \includegraphics[width=.5\columnwidth]{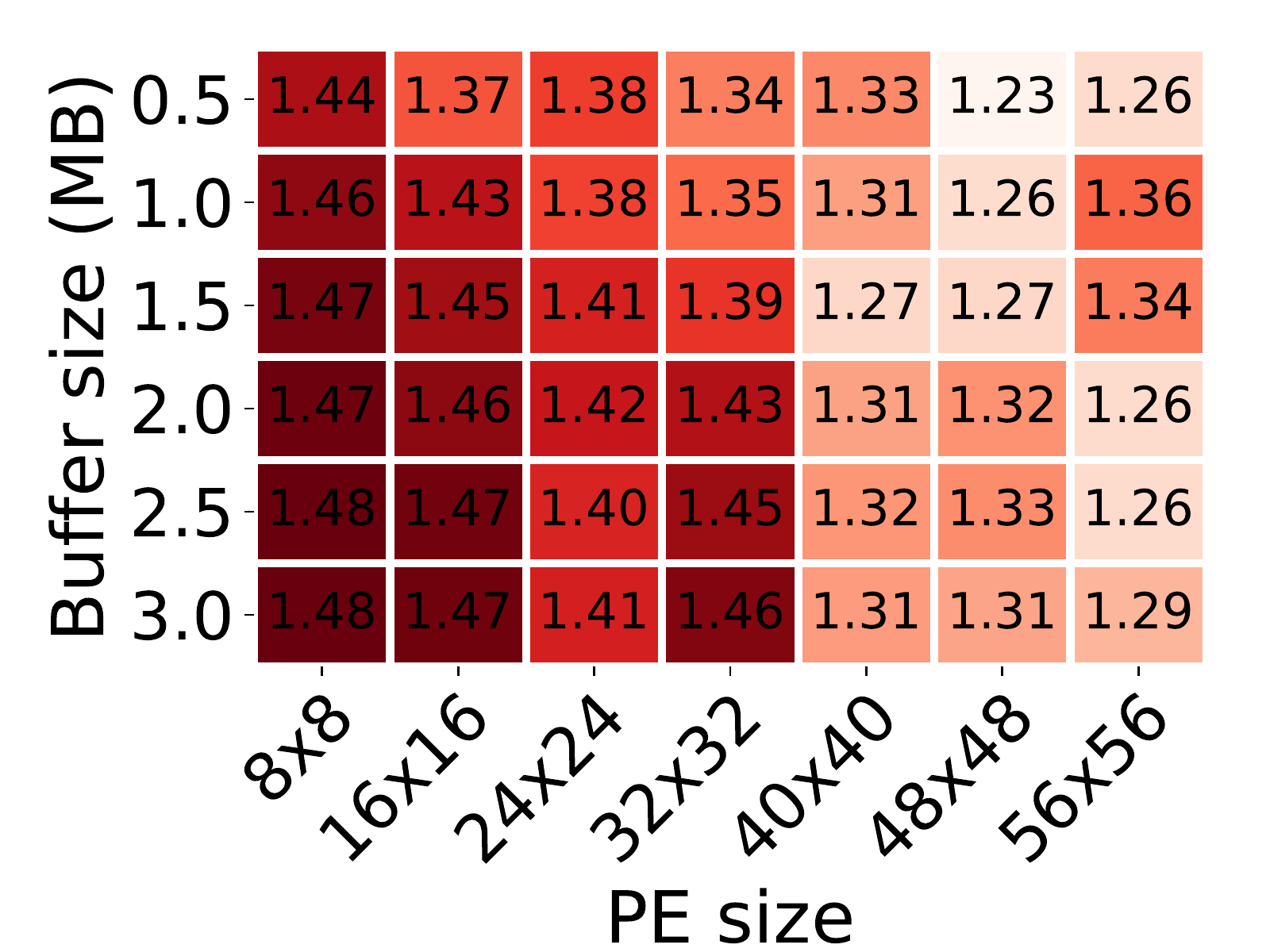}
  \label{fig:sensitivity_speedup_2d}
  }
  \subfloat[\small{Energy reduction.}]
  {
  \includegraphics[width=.5\columnwidth]{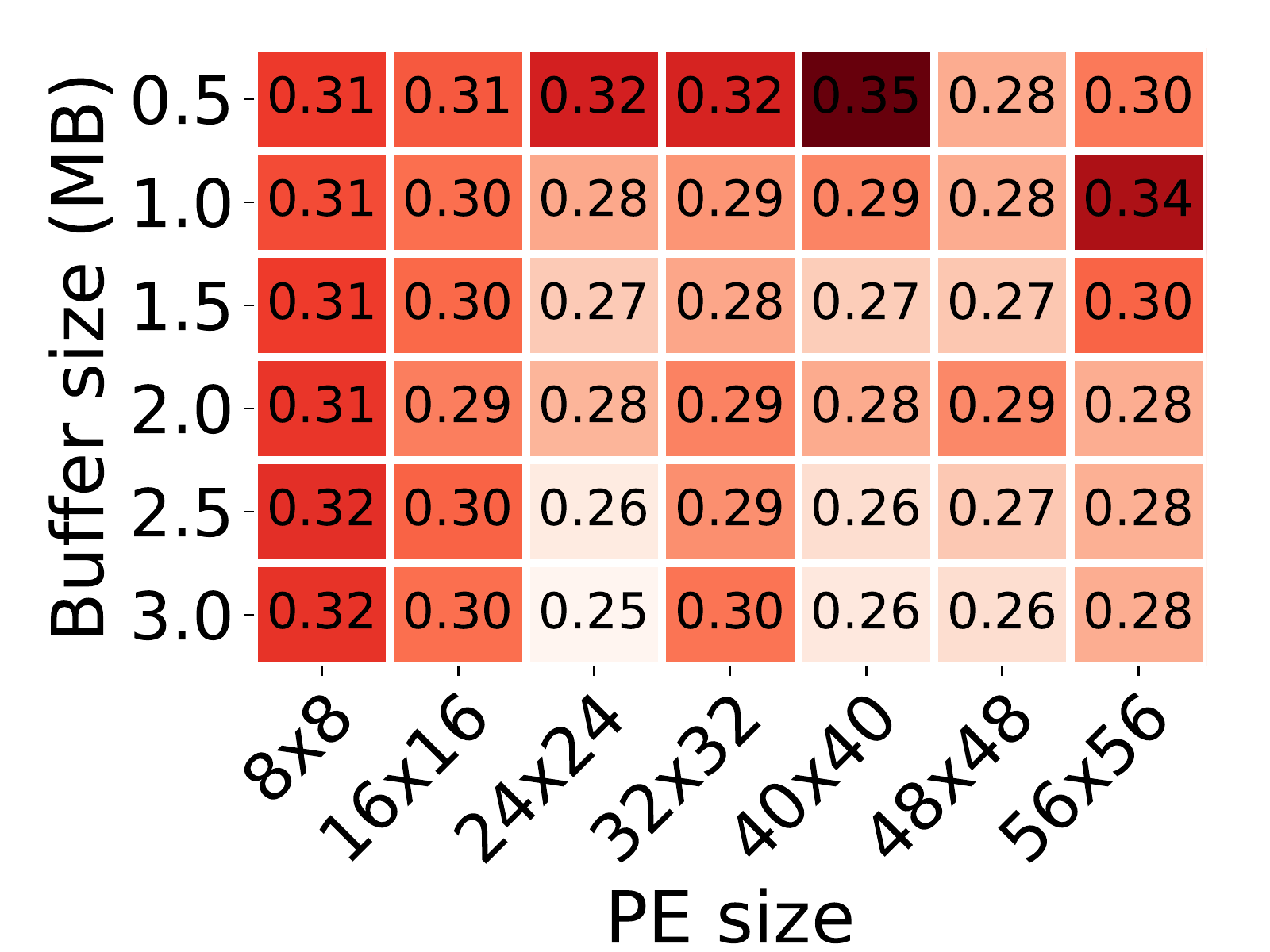}
  \label{fig:sensitivity_energy_2d}
  }
  \caption{Sensitivity analysis of DCO speedup and energy reduction with buffer size and PE array size on \textsc{FlowNetC}. Speedup is normalized to the corresponding configurations, not to a single, common baseline.}
  \label{fig:sensitivity_2d}
\end{figure}

While the speedup and energy reduction studied so far are based on one representative baseline accelerator configuration (\Sect{sec:exp:setup}), we find that our deconvolution optimization generally achieves similar improvements on other hardware configurations with resource provisions. In particular, we focus on two key types of hardware resource: PE size and on-chip buffer size. For brevity we only report results on \textsc{FlowNetC}~\cite{FlownetC2015}, but the trends generally hold.

\Fig{fig:sensitivity_speedup_2d} and \Fig{fig:sensitivity_energy_2d} show how DCO's average speedup and energy reduction of the entire network  vary with different PE size and buffer size combinations, respectively. Note that the results are normalized to their corresponding hardware configurations rather than to the baseline described in \Sect{sec:exp:setup}. For instance, on the hardware with an 8$\times$8 PE array and a 0.5~MB on-chip buffer, DCO achieves an 1.44$\times$ speedup.

DCO achieves speedups of 1.2$\times$ -- 1.5$\times$ and energy reductions of 25\% -- 35\% across different hardware capabilities, demonstrating broad applicability. 
In general, the performance improvement of DCO is more pronounced with small PE arrays, where the performance is compute-bound. As the PE size increases, the performance becomes memory bound, such that memory bandwidth limitations mask the benefit of data reuse. In addition, as the buffer size increases, the reuse opportunities inherently exposed by the buffer is higher, and thus data reuse optimizations become less critical, hence the lower energy savings.


\subsection{Eyeriss and GPU Comparisons}
\label{sec:eval:eyeriss}

\begin{figure}[t]
  \centering
  \captionsetup[subfigure]{width=0.5\columnwidth}
  \subfloat[\small{Speedup.}]
  {
  \includegraphics[width=.5\columnwidth]{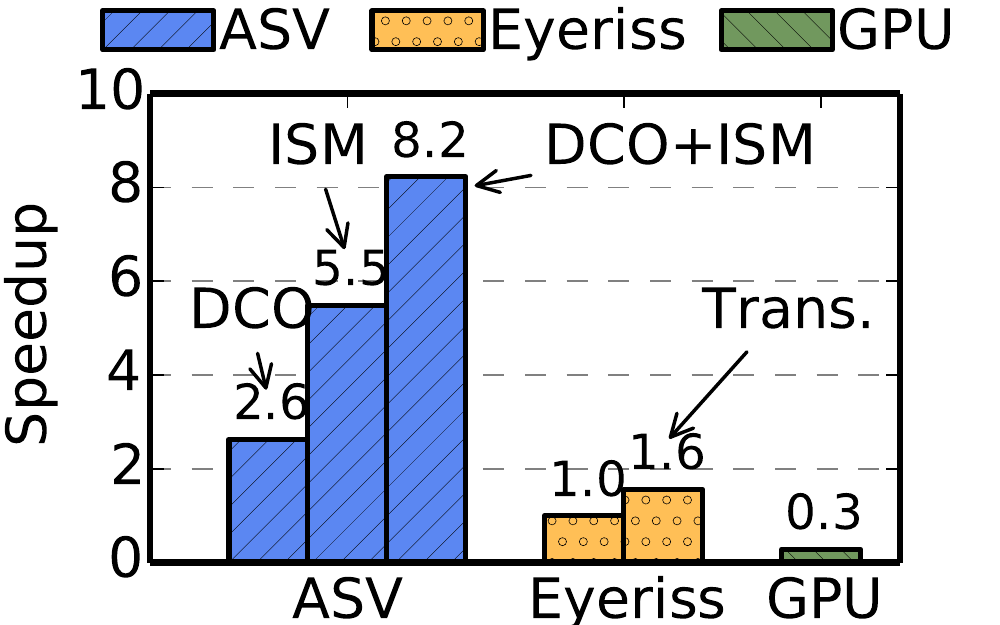}
  \label{fig:overall_speedup}
  }
  \subfloat[\small{Norm. energy. Lower is better.}]
  {
  \includegraphics[width=.5\columnwidth]{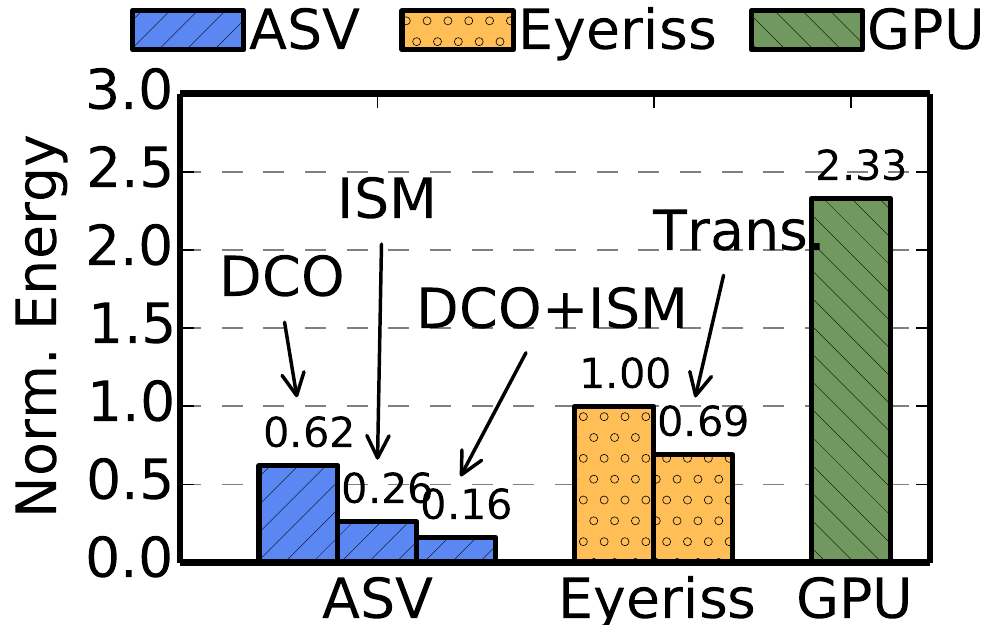}
  \label{fig:overall_energy}
  }
  \caption{Comparison of speedup and energy reduction for \proj, Eyeriss, and GPU. The results are normalized to Eyeriss. We show three variants of \proj. We also apply the deconvolution transformation to Eyeriss to obtain a stronger baseline, hence the \texttt{DCT} bar of Eyeriss.}
  \label{fig:overall_result}
\end{figure}

For completeness, we also compare \proj with Eyeriss~\cite{Eyeriss2016} and a mobile Pascal GPU~\cite{tx2dev}.~\Fig{fig:overall_result} shows the speedup and energy reductions of the three \proj variants, Eyeriss and GPU. Data is normalized to Eyeriss. To obtain a stronger Eyeriss baseline, we extended the Eyeriss simulator~\cite{eyerisssim, gao2017tetris} to support our deconvolution optimization. Our ILAR optimization does not apply because Eyeriss's spatial architecture requires a different reuse formulation from the one presented here, which targets a systolic array.

On average, when combing DCO and ISM, \proj achieves $8.2\times$ speedup against Eyeriss while consuming only $16\%$ of the energy. DCO and ISM contribute to $38\%$ and $74\%$ on energy saving, respectively. Critically, Eyeriss can also benefit from the deconvolution transformation (\texttt{DCT}), which achieves a $ 1.6\times $ speedup and 31\% energy saving compared to the baseline Eyeriss. Finally, \proj is 27$\times$ faster while consuming 15$\times$ lower energy than the GPU.

\subsection{GANNX Comparison}
\label{sec:eval:gannx}

\begin{figure}[t]
  \centering
  \captionsetup[subfigure]{width=0.5\columnwidth}
  \subfloat[\small{Speedup.}]
  {
  \includegraphics[width=.48\columnwidth]{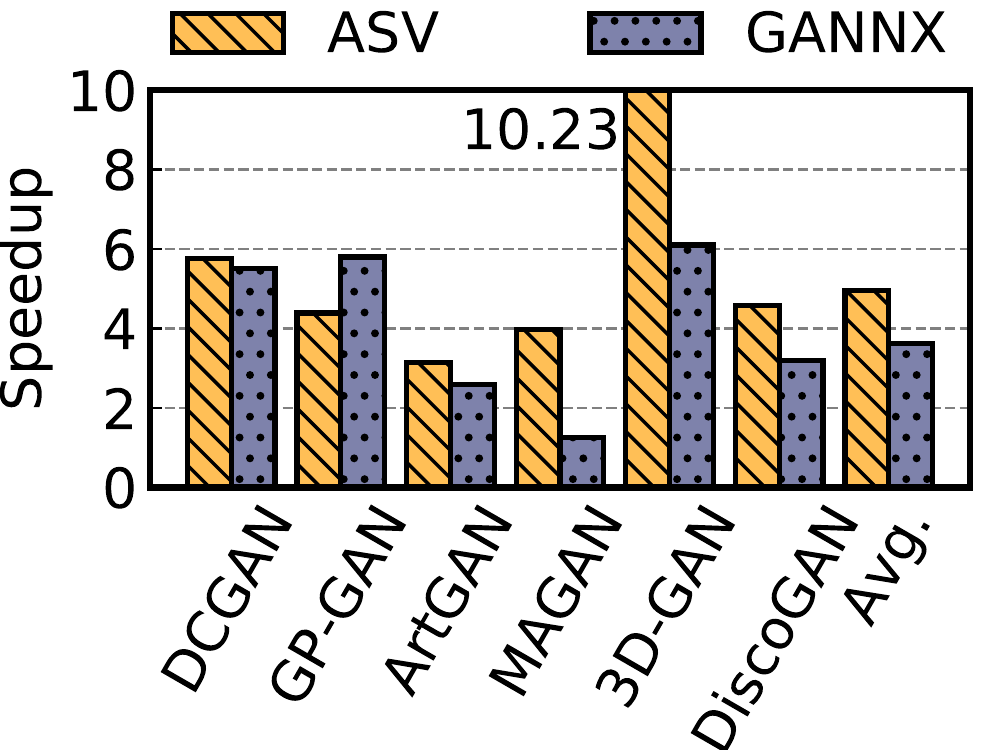}
  \label{fig:gannx_speedup}
  }
  \subfloat[\small{Energy reduction.}]
  {
  \includegraphics[width=.48\columnwidth]{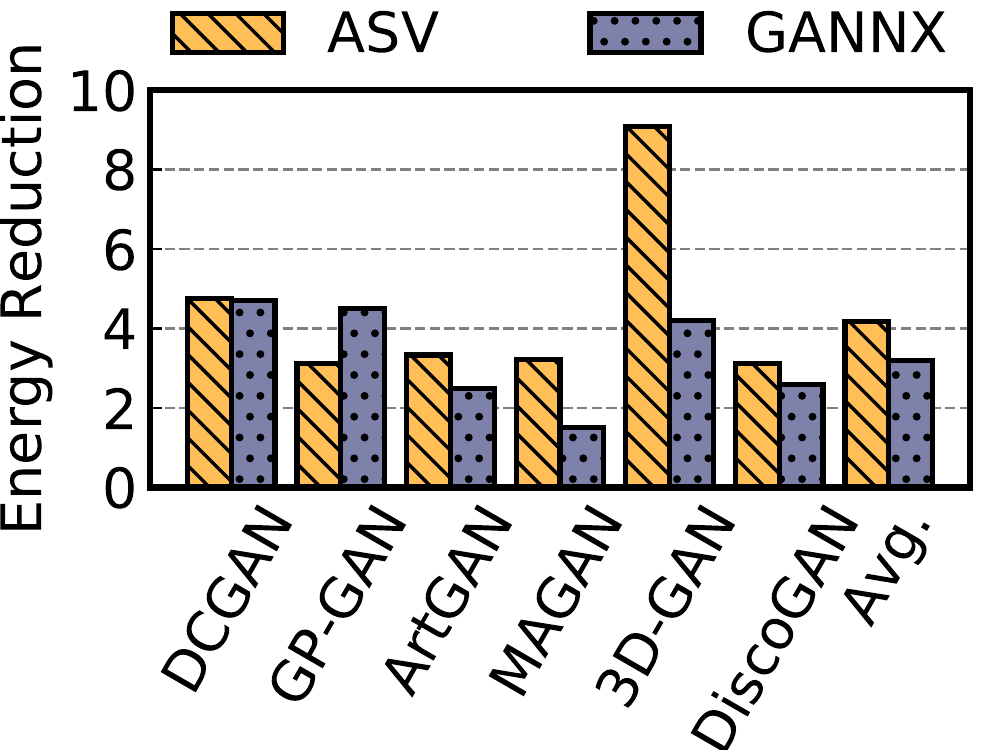}
  \label{fig:gannx_energy}
  }
  \caption{Speedups and energy reductions on GANs between \proj and GANNX. Results are normalized to Eyeriss.}
  \label{fig:gannx_result}
\end{figure}

Our deconvolution optimizations are not limited to stereo DNNs, but also apply broadly to deconvolution. To demonstrate general applicability, we apply the deconvolution optimizations to Generative Adversarial Networks (GANs), a class of DNNs that also make heavy use of deconvolution layers~\cite{goodfellow2014generative}. We compare against GANNX~\cite{Yazdanbakhsh2018GAN}, a dedicated DNN accelerator for accelerating deconvolution in GANs. We configure both \proj and GANNX to have the same PE and buffer sizes. We normalize the \proj results to Eyeriss, consistent with what GANNX reports.

\Fig{fig:gannx_result} shows speedup and energy comparisons across the six GANs used by GANNX. On average, \proj achieves $5.0\times$ speedup and $4.2\times$ energy reduction, higher than the improvements from GANNX ($ 3.6\times $ speedup and $3.2\times$ energy reduction). The higher gains from \proj are mainly attributed to the inter-layer activation reuse, which is uniquely exposed by our deconvolution transformation and is unavailable in GANNX. Critically, our deconvolution optimizations are purely software optimizations without requiring the specialized hardware support of GANNX.

\section{Related Work}
\label{sec:related}

\paragraph{Stereo Vision Accelerators} Recently, commercial mobile vision systems~\cite{Zhu2018mlsoc} have started integrating dedicated stereo accelerators, such as the Stereo Depth Block in the Movidius Enhanced Vision Accelerator Suite~\cite{myriadx}, and the Stereo \& Optical Flow Engine (SOFE) in the Nvidia Xavier mobile SoC~\cite{xaviersoc}. From publicly available details, these are fixed-functioned accelerators targeting classic stereo algorithms, similar to previous stereo vision accelerators~\cite{mazumdar2017exploring, wang2015real_time_stereo_system,  gudis2013stereoembedded, ttofis2014high, yang2014hardware}. In contrast, \proj combines the efficiency of classic stereo algorithms with the accuracy of stereo DNNs.


\paragraph{Motion-based Algorithms} Our ISM algorithm shares a similar key observation as some recent motion-based vision algorithms such as EVA$^2$~\cite{buckler2018eva2} and Euphrates~\cite{zhu2018euphrates}, in that correlation across frames in a video stream
can be used to simplify continuous vision tasks. Euphrates~\cite{zhu2018euphrates} focuses on computing regions-of-interest (ROIs) in object detection and tracking tasks. In contrast, stereo vision is concerned with the depth of the whole frame rather than discrete ROIs. 
EVA$^2$~\cite{buckler2018eva2} is not limited to ROIs. However, it relies on estimating the motion of an intermediate activation's receptive field. In the stereo task, the receptive field of an intermediate activation necessarily spans both the left and right images. Thus, the motion of the receptive field would be difficult, if not impossible, to calculate.


Fundamentally, motion-based relaxations fall within the realm of incremental computing, a general technique used in program analysis
and optimization~\cite{michie1968memo, pugh1989incremental} and applies beyond the temporal and/or vision domain. 
Diffy~\cite{mahmoud2018diffy} exploits the spatial similarity across pixels in the same frame to improve DNN efficiency. Riera et al.~\cite{riera2018computation} exploit repeated \textit{ifmap} elements in speech recognition.

\paragraph{Deconvolution Sparsity} Many prior studies optimize hardware to exploit sparsity in DNNs~\cite{shiftnet2018, wen2016learning_Sparsity, He2017channel_pruning, kung2018packing, Parashar2017scnn, lee2019sparse, whatmough2018sparse, Cnvlutin2016, EIE2016}. 
Stereo vision DNNs make use of deconvolution layers, which expose structured sparsity patterns. 
Recent work has prosed specialized hardware specifically for exploiting sparsity in deconvolution layers~\cite{Yazdanbakhsh2018GAN, song2018GAN}. Our observation, however, is that mitigating sparsity-induced efficiencies in deconvolution does not necessarily require hardware support. We propose novel software optimizations to eliminate the compute inefficiencies without hardware changes.



\paragraph{Data-Reuse Optimization} Exploiting data-reuse (through tiling) is critical to DNN efficiency~\cite{Eyeriss2016, yang2016systematic, gao2017tetris, gao2019tangram, DaDianNao2014, Ma2017dnnFPGA, Hegde2018morph, Mullapudi2016AutoHalide, yang2018dnn, whatmough2019vlsi, FusedCNN2016, lu2017flexflow, kwon2018maeri, whatmough2019fixy}. Orthogonal to generic data-reuse, we identify a new reuse dimension, inter-layer activation reuse (\texttt{ILAR}), that is uniquely enabled by our deconvolution transformation.


Previous DNN mapping frameworks mostly rely on exhaustive search~\cite{yang2018dnn, yang2016systematic, Hegde2018morph}, which does not scale to exploiting \texttt{ILAR} (\Sect{sec:deconv:reuse}). 
Instead, \proj uses a constrained-optimization that can solved efficiently using dynamic programming. 
TETRIS~\cite{gao2017tetris} also uses a constrained-optimization for DNN scheduling, albeit with certain problem-specific simplifications. 
However, it does not exploit \texttt{ILAR}. Our formulation directly optimizes for latency rather than memory traffic~\cite{yang2016systematic, gao2017tetris} or resource utilization~\cite{Ma2017dnnFPGA}.


\section{Conclusion} 
\label{sec:conc}


\proj simultaneously improves performance and energy-efficiency of ``depth from stereo'', while maintaining high accuracy. 
\proj combines algorithmic and computational optimizations that leverage characteristics unique to stereo vision. We demonstrate careful design choices that let these optimizations be integrated with existing DNN accelerators with minor hardware extensions. As intelligent machine perception increasingly relies on depth sensing, \proj provides a promising first step towards comprehensive system support.
\begin{appendices}

\section{General Deconvolution Transformation with an N-dimensional Kernel}
\label{sec:appendix:decompose}
Here we show a general formulation for decomposing a deconvolution kernel. A N-dimension kernel is decomposed into $2^N$ sub-kernels, each sub-kernel $S_{k}$ can be calculated as follows:

$$ S_{k}^{(i_{0}, i_{1},..., i_{N-1})} = K^{(2i_{0}+\delta_{0}, 2i_{1}+\delta_{1},..., 2i_{N-1} + \delta_{N-1})},~~~\forall k \in [0, 2^N-1] $$
$$ \delta_{j} = (k \gg j~\& 1),i_{j} \in [0, \lfloor(|K^{(j)}|-\delta_{j})/2\rfloor),~~~\forall j \in [0, N-1] $$

where N is the number of dimensions in the original kernel $K$, $S_{k}^{(i_{0}, i_{1},..., i_{N-1})}$ is the element $(i_{0}, i_{1},..., i_{N-1})$ in $k^{th}$ sub-kernel, and $K^{(*,...,*)}$ denotes an element in the original kernel $K$.

Our formulation essentially shows that the element $(i_{0}, i_{1},..., i_{N-1})$ in the $k^{th}$ sub-kernel comes from the element $(2i_{0}+\delta_{0}, 2i_{1}+\delta_{1},..., 2i_{N-1} + \delta_{N-1})$ in the original kernel. Each $\delta_*$ is a binary value calculated by: $\delta_{i} = (k \gg j \& 1)$, where $\gg$ is the right shift operator and $\&$ is the bitwise AND operator. The dimension of the each sub-kernel is determined by: $i_{j} \in [0, \lfloor(|K^{(j)}|-\delta_{j})/2\rfloor)$, where $|K^{(j)}|$ is the size of $j$th dimension of the original kernel.

\section{The Expressions of $\Delta W^i_k$, $\Delta IF^i$, and $\Delta OF^i_k$}
\label{sec:appendix:exp}
The expression of $\Delta W^i_k$, $\Delta IF^i$, and $\Delta OF^i_k$ are determined by $W^i_k$, $H^i_k$, and $|\overrightarrow{C^i}|$. Their exact expressions are shown below:

$$ \Delta W^i_k = W_k^i \times H_k^i \times C_k^i$$

\noindent where ${C_k^i}$ denotes the total number of sub-kernel $k$ in round $i$; $W^i_k$ and $H^i_k$ are the dimensions of sub-kernel $k$ in round $i$. 

$$\Delta IF^i = W^i \times H^i \times I$$ 

\noindent where $W^i$ and $H^i$ are the weight and height of an \textit{ifmap} tile to be loaded in round $i$, and $I$ is the number of input channels.
$$\Delta OF^i_k = \frac{W^i \times H^i \times C_k^i}{s^2}$$

\noindent where $s$ denotes the stride of this layer.
\end{appendices}


\raggedright
\balance
\bibliographystyle{ACM-Reference-Format}
\bibliography{refs}
\end{document}